\def\year{2021}\relax
\documentclass[letterpaper]{article} 
\usepackage{aaai21}  
\usepackage{times}  
\usepackage{helvet} 
\usepackage{courier}  
\usepackage[hyphens]{url}  
\usepackage{graphicx} 
\usepackage{natbib}  
\usepackage{caption} 
\usepackage{marvosym}
\usepackage{amsmath,amssymb,amsfonts}
\usepackage{xcolor}
\usepackage{pifont}
\usepackage{wrapfig}
\usepackage{caption}
\usepackage{subfigure}
\usepackage{graphicx}
\usepackage{subfigure} 
\usepackage{eqparbox,array} 
\usepackage{amsmath}
\usepackage{amsfonts,amssymb}
\usepackage{algorithmic}
\usepackage{algorithm}  
\usepackage{algorithmic,eqparbox,array} 
\usepackage{indentfirst}
\usepackage{verbatim}
\usepackage{float}
\usepackage{color}
\usepackage{threeparttable}
\usepackage{appendix}
\usepackage{multirow}
\UseRawInputEncoding
\newtheorem{theorem}{Theorem}

\newtheorem{lemma}{Lemma}
\newtheorem{proposition}{Proposition}
\newtheorem{corollary}{Corollary}
\newtheorem{definition}{Definition}

\def\BibTeX{{\rm B\kern-.05em{\sc i\kern-.025em b}\kern-.08em
		T\kern-.1667em\lower.7ex\hbox{E}\kern-.125emX}}

\title{FLAME: Differentially Private Federated Learning in the Shuffle Model}
\author{
	Ruixuan Liu\textsuperscript{\rm 1}, 
	Yang Cao\textsuperscript{\rm 2}\thanks{Corresponding authors: Hong Chen, Yang Cao},
	Hong Chen\textsuperscript{\rm 1*}, 
	Ruoyang Guo\textsuperscript{\rm 1}, 
	Masatoshi Yoshikawa\textsuperscript{\rm 2}\\ 
	}
\affiliations{
	\textsuperscript{\rm 1}Renmin University of China\\
	\textsuperscript{\rm 2}Kyoto University\\
    ruixuan.liu@ruc.edu.cn,
    yang@i.kyoto-u.ac.jp,
    chong@ruc.edu.cn, 
    ryguo@ruc.edu.cn,
    yoshikawa@i.kyoto-u.ac.jp
	}

\begin{document}
\maketitle

\begin{abstract}
	Federated Learning (FL) is a promising machine learning paradigm that enables the analyzer to train a model without collecting users' raw data. 
	To ensure users' privacy, differentially private federated learning has been intensively studied.
	The existing works are mainly based on the \textit{curator model} or \textit{local model} of differential privacy.
	However, both of them have pros and cons.
	The curator model allows greater accuracy but requires a trusted analyzer.  
	In the local model where users randomize local data before sending them to the analyzer, a trusted analyzer is not required but the accuracy is limited.
	In this work, by leveraging the \textit{privacy amplification} effect in the recently proposed shuffle model of differential privacy, we achieve 
	the best of two worlds, i.e., accuracy in the curator model and strong privacy without relying on any trusted party.
	We first propose an FL framework in the shuffle model and a simple protocol (SS-Simple) extended from existing work.
	We find that SS-Simple only provides an insufficient privacy amplification effect in FL since the dimension of the model parameter is quite large. 
	To solve this challenge, we propose an enhanced protocol (SS-Double) to increase the privacy amplification effect by subsampling.
	Furthermore, for boosting the utility when the model size is greater than the user population, we propose an advanced protocol (SS-Topk) with gradient sparsification techniques.
	We also provide theoretical analysis and numerical evaluations of the privacy amplification of the proposed protocols.
	Experiments on real-world dataset validate that SS-Topk improves the testing accuracy by 60.7\% than the local model based FL. 
	We highlight an observation that SS-Topk improves the accuracy by 33.94\% than the curator model based FL without any trusted party.
	Compared with non-private FL, our protocol SS-Topk only lose 1.48\% accuracy under $(2.348, 5e^{-6})$-DP per epoch. 
\end{abstract}
\section{Introduction}
\noindent
Federated Learning (FL) \cite{mcmahan2016Federated} is a promising machine learning paradigm that enables the analyzer to train a central model by collecting users' local updates instead of the raw data.
However, it has been shown that sharing raw local updates compromises users' privacy \cite{nasr2019comprehensive,hitaj2017deep,zhu2019deep}.
To this end, \textit{differentially private federated learning} has been wildly studied to provide formal privacy.
\begin{figure*}[t]
	\centering
	\includegraphics[width=0.85\textwidth,trim=0 213 0 100,clip]{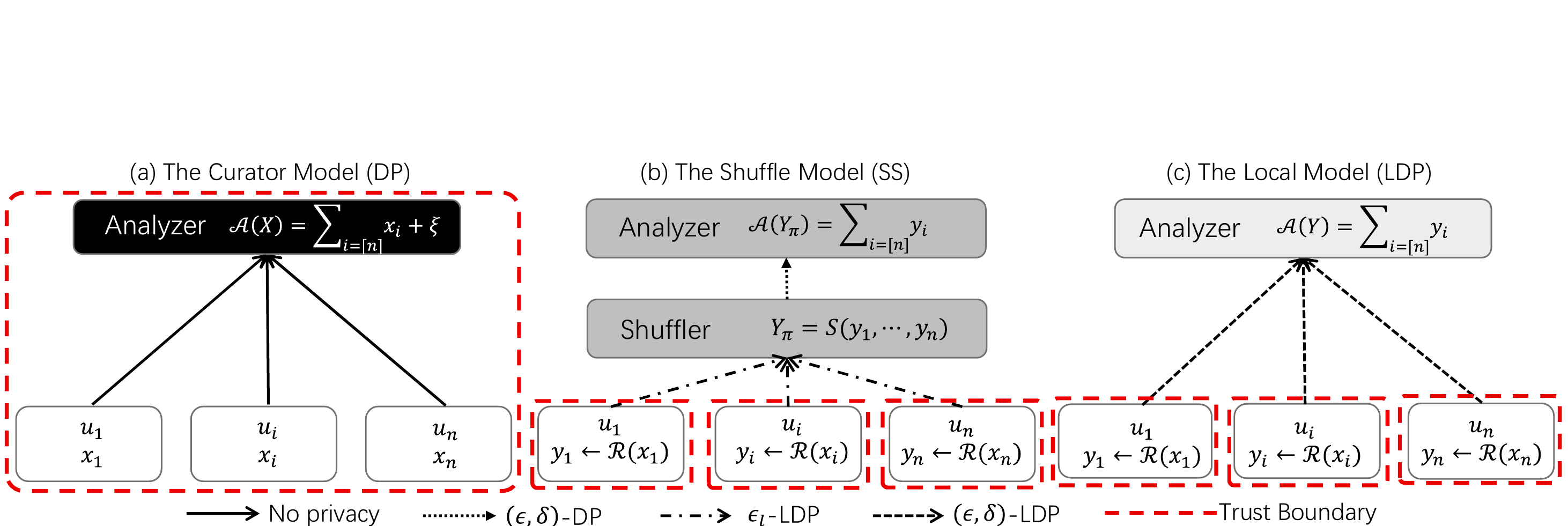}
	\caption{
		Trust boundaries in the curator, shuffle and local model of differential privacy.
		The curator model relies on a trusted analyzer.
		Users in the local model or the shuffle model do not need to trust any external parties and thus preserve strong privacy.
		}\label{fig-3trust}
\end{figure*}
The existing works are mainly based on the \textit{curator model} (DP) or \textit{local model} (LDP) of differential privacy.
The curator model based FL (DP-FL) \cite{mcmahan_learning_2018,geyer2017differentially} allows better learning accuracy but relies on a trusted analyzer to collect raw local updates.
The local model based FL (LDP-FL) \cite{wang2019collecting, liu2020FedSel} preserves strong local privacy since the users randomize local updates before sending them to an untrusted analyzer; but it suffers a low utility.
Specifically, for a basic bit summation task over $n$ users with privacy budget $\epsilon$\footnote[1]{A larger privacy budget leads to better utility and less privacy.}, the error of DP can achieve $O(1/\epsilon)$;
whereas, the error of LDP is bounded by $\Omega(\sqrt{n})$ \cite{chan2012optimal}.

The recently proposed secure shuffle model (SS) can achieve the best of two worlds, i.e., accuracy in the curator model and strong privacy in the local model.
The shuffle model introduces a shuffler (sitting between users and the analyzer, as shown in Figure \ref{fig-3trust}(b)) to permute the locally randomized data before users transmitting them.
The accuracy gain of the shuffle model is obtained from the \textit{privacy amplification} effect \cite{erlingsson_amplification_2018}, which indicates that the shuffled  (i.e., anonymized) outputs of local randomizers provide a stronger (amplified) privacy in the central view of differential privacy than the one without a shuffler.
Accordingly, less local noise is needed in the shuffle model for the same level of privacy against the untrusted analyzer.

However, it is not clear how to employ the shuffle model in federated learning.
Although a few works have investigated basic tasks such as the bit/real summation \cite{balle_privacy_2019} and histogram \cite{balcer2019separating}, the existing protocols may not be viable for the multi-dimensional aggregation FL.
This is because the error brought by local noises can be aggravated with a dimensional factor \cite{duchi2013local}.
Moreover, since the number of users participating in one iteration is typically a few thousands, the aggregation escalates into a high-dimensional task. 
We solve the above challenges by making the following contributions:
\begin{itemize}
	\setlength{\itemsep}{0pt}
	\setlength{\parsep}{0pt}
	\setlength{\parskip}{0pt}
	\item 
	For the first time, we propose FLAME, a federated learning framework in the shuffle model such that the users enjoy strong privacy and the analyzer enjoys the accuracy of the model.
	We first formalize our privacy goal in FLAME by clarifying the trust boundary and fine-grained trust separation (Table \ref{tab-privacy}), then we propose SS-Simple protocol by extending a one-dimensional task \cite{balle_privacy_2019}.
	We find that, although the privacy amplification is achievable by SS-Simple, the magnitude of amplification diminishes with the dimension size of the local updates (Corollary \ref{coro-simple-l-c}).
	\item 
	To alleviate this challenge, we propose SS-Double protocol to enhance the privacy amplification by subsampling.
	As we notice that the amplification by subsampling may not be composable with shuffling, we propose a novel \textit{dummy padding} method to bridge two kinds of amplification effects with formal proofs (Theorems \ref{theo-double-lk-cd} and \ref{theo-double-cd-c}).
	We demonstrate that the SS-Double enjoys \textbf{dozens} times privacy amplification comparing with SS-Simple (Figure \ref{fig-krr-bound}).
	\item
	A problem of SS-Double protocol is that the random subsampling treats all dimensions equally and thus may discard ``important'' dimensions.
	To further boost the utility in a high-dimensional case, we design an advanced protocol called SS-Topk, which is based on the idea of gradient sparsification.
	A challenge is that the indexes of Top-$k$ elements in the local update vector may reveal sensitive information to the shuffler since the selection is data-dependent.
	We quantify this privacy threat by formalizing \textit{index privacy} and design a method to flexibly trade off between index privacy and utility.
	We note that the index privacy will not impair the privacy against the analyzer.
	\item Finally, we conduct experiments on the real-world dataset to validate the effectiveness of the proposed protocols. 
	It turns out that the proposed double amplification effect in SS-Double and private dimension selection in SS-Topk significantly improve the learning accuracy. 
	We observe a 33.94\% accuracy improvement of SS-Topk than the curator model based FL without relying on any trusted party.
	Compared with non-private FL, SS-Topk only lose 1.48\% accuracy under $(2.348, 5e^{-6})$-DP per epoch.
\end{itemize}
\section{Preliminaries}
\noindent
We first clarify the trust boundaries in Figure \ref{fig-3trust} and then introduce properties that will be used throughout the paper and formally define our problem.

\subsection{The Curator and Local Models}
\noindent
In the curator model, a trusted analyzer collects users' raw data (e.g local updates) and executes a private mechanism for differentially private outputs.
The privacy goal is to achieve indistinguishability for any outputs w.r.t. two neighboring datasets which differ by replacing one user's data, denoted as $X \simeq_r X^\prime$.
We have the following definition:
\begin{definition}\label{def_cdp}[Differential Privacy (DP)]
	A mechanism $\mathcal{M}:\mathbb{X}^n \rightarrow \mathbb{Z}$ satisfies $(\epsilon, \delta)$-differentially privacy if for any two neighboring datasets $X \simeq_r X^{\prime}\in\mathbb{X}^n$ and any subsets $S \subseteq \mathbb{Z}$,
	$
	\Pr[\mathcal{M}(X) \in S]\le e^\epsilon \Pr[\mathcal{M}(X^\prime)\in S] + \delta.
	$
\end{definition}

However, the curator model assumes the availability of a trusted analyzer to collect raw data.
Local differential privacy
in Definition \ref{def_ldp} does not rely on any trusted party because users send randomized data to the server.
If $\mathcal{R} $ satisfies $(\epsilon, \delta)$-LDP, observing collected results $( y_1, \cdots, y_n )$ or the summation implies $(\epsilon, \delta)$-DP \cite{dwork2014the(book)}.

\begin{definition}\label{def_ldp}[Local Differential Privacy (LDP)]
	\setlength{\itemsep}{0pt}
	\setlength{\parsep}{0pt}
	\setlength{\parskip}{0pt}
	A mechanism $\mathcal{R}:\mathbb{X}\rightarrow \mathbb{Y}$ satisfies $(\epsilon, \delta)$-locally differentially private if for any two inputs $x, x^\prime \in\mathbb{X}$ and any output $y \in \mathbb{Y}$,
	$
	\Pr[\mathcal{R}(x) = y] \le e^\epsilon \Pr[\mathcal{R}(x^\prime) = y] + \delta.
	$
\end{definition}

\subsection{The Shuffle Model}
\noindent
The protocol of a shuffle model consists of three components:  $\mathcal{P}=\mathcal{A}\circ\mathcal{S}\circ\mathcal{R}^n$, as shown in Figure \ref{fig-3trust}(b).
Existing works \cite{balle_differentially_2019,balcer2019separating,cheu_distributed_2019,ghazi_power_2020,ghazi2020pure} focus on the basic task where each user holds a one-dimensional data $x\in\mathbb{X}$.
We denote $n$ users' data as the dataset $X=(x_1, \cdots, x_n)\in \mathbb{X}^n$.
Each user runs a randomizer $\mathcal{R}:\mathbb{X}\rightarrow \mathbb{Y}^{m}$ to perturb the local data into $m$ messages that satisfy $\epsilon_l$-LDP. 
W.o.l.g. we focus on the \textit{single-message} protocol where $m=1$.
The shuffler executes $\mathcal{S}: \mathbb{Y}^* \rightarrow \mathbb{Y}^*$ with a uniformly random permutation $\pi$ over received messages.
The analyzing function $\mathcal{A}:\mathbb{Y}^*\rightarrow \mathbb{Z}$ takes the shuffled messages as input and outputs the analyzing result.

The privacy goal in shuffle model is to ensure $\mathcal{M}=\mathcal{S}\circ \mathcal{R}^n$ satisfies $(\epsilon_c,\delta_c)$-DP, because $\mathcal{A}$ is executed by an untrusted analyzer, who is not obliged to protect users' privacy.
By the post-processing property \cite{dwork2014the(book)}, the protocol $\mathcal{P}$ achieves the same privacy level as $\mathcal{M}$. 
Hence, we focus on analyzing the indistinguishability for $\mathcal{M}(X)$ and $\mathcal{M}(X^\prime)$.
\cite{erlingsson_amplification_2018} proved that the privacy of $\mathcal{M}$ can be ``amplified''.
In other words, when each user applies the local privacy budget $\epsilon_l$ in $\mathcal{R}$, $\mathcal{M}$ can achieve a stronger privacy of $(\epsilon_c, \delta_c )$-DP with $\epsilon_c < \epsilon_l$.
Compared with the local model, the shuffle model needs less noise to achieve the same privacy level.

Among existing works, the privacy blanket \cite{balle_privacy_2019} provides an optimal amplification bound for the \textit{single-message} protocol.
The analyzing intuition is to linearly decompose the output distribution into a data-dependent distribution and a uniform random ``privacy blanket" distribution.
With the local randomizer $\mathcal{R}_{\gamma, b}$ in Algorithm \ref{alg-krr-randomizer}, where $\gamma=\frac{b}{e^{\epsilon_l}+b-1}$ denotes the probability to output an element from the blanket distribution.
the input value $x$ is encoded into a discrete domain $[b]$ and then randomized.
After $\mathcal{S}$ runs a permutation, $\mathcal{A}$ aggregates shuffled results with $\hat{z} \leftarrow \frac{1}{b} \sum_{i=i}^n y_i$ and de-bias with
\begin{equation}\label{eq-debias}
z\leftarrow (\hat{z}-n\gamma/2)/(1-\gamma).
\end{equation}
The privacy amplification bound for Algorithm \ref{alg-krr-randomizer} is shown in Lemma \ref{lemma-krr-bound} and the effect for generic randomizer is distilled in Corollary \ref{corollary5.3.1}.
For a randomizer with Laplace Mechanism on the domain $[0, 1]$, $\gamma=e^{-\epsilon_l/2}$.
A tighter bound (i.e., a greater amplification) can be accessed with numerical evaluations.
\begin{algorithm}[t]
	\small
	\caption{$\mathcal{R}_{\gamma, b}:[0, 1]\rightarrow [b]$ \cite{balle_privacy_2019}}\label{alg-krr-randomizer}
	\begin{algorithmic}[1]
		\REQUIRE input scalar $x\in [0, 1]$ 
		\ENSURE perturbed value $y\in[b]$
		\STATE $\bar{x} \leftarrow \lfloor xb \rfloor + \text{Ber}(xb-\lfloor xb \rfloor)$
		\STATE Sample $r \leftarrow Ber(\gamma)$
		\STATE
		$
		y=
		\begin{cases}
		\bar{x} & \text{if } r=0,\\
		\text{Unif}(\{1, \cdots, b\}) & \text{else}.
		\end{cases}
		$
	\end{algorithmic}
\end{algorithm}

\begin{lemma}\label{lemma-krr-bound} \cite{balle_privacy_2019}
	For $\sqrt{\frac{14\log(2/\delta_c)(b-1)}{n-1}} < \epsilon_c \le 1$, if $\mathcal{R}_{\gamma, b}$ satisfies $\epsilon_l$-LDP, we have $(\epsilon_c, \delta_c)$ for $\mathcal{S}\circ \mathcal{R}^n$, where $\epsilon_c= \sqrt{\frac{14\log(2/\delta_c) (e^{\epsilon_l}+b-1)}{n-1}}$.
\end{lemma}
\begin{corollary}\label{corollary5.3.1}  \cite{balle_privacy_2019}
	\setlength{\itemsep}{0pt}
	\setlength{\parsep}{0pt}
	\setlength{\parskip}{0pt}
	In the shuffle model, if $\mathcal{R}$ is $\epsilon_l$-LDP, where $\epsilon_l\le \log (n/log(1/\delta_c))/2$.
	$\mathcal{M}$ satisfies $(\epsilon_c, \delta_c)$-DP with:
	$\epsilon_c=O((1 \land \epsilon_l) e^{\epsilon_l} \sqrt{\log (1/ \delta_c)/n} )$.
\end{corollary} 

\subsection{Composition and Subsampling Properties}
\noindent
The composition properties \cite{dwork2010boosting} are generic for both the curator and the local model of differential privacy.
\begin{lemma}\label{lemma-sequential-composition}
	$\forall \epsilon \ge 0, t \in\mathbb{N}$,
	the family of $\epsilon$-DP mechanism satisfies $t\epsilon$-DP under $t$-fold adaptive composition.
\end{lemma}

\begin{lemma}\label{lemma-advanced-composition}
	$\forall \epsilon, \delta, \delta^\prime>0, t\in\mathbb{N}$, the family of $(\epsilon, \delta)$-DP mechanism satisfies $(\sqrt{2t\ln(1/\delta^\prime)} \cdot \epsilon + t\cdot \epsilon(e^\epsilon-1), t\delta+\delta^\prime)$-DP under $t$-fold adaptive composition.
\end{lemma}

A mechanism $\mathcal{K}$ that randomly subsamples $m$ elements \textit{without replacement} from a database with $n$ records leads to a \textit{privacy amplification} by subsampling in Lemma \ref{lemma-subsampling}.
\begin{lemma}\label{lemma-subsampling}[Privacy Amplification by Subsampling] \cite{balle_privacy_2018}
	If {\small $\mathcal{M}:\mathbb{X}^m\to \mathbb{Y}$} satisfies $(\epsilon, \delta)$-DP with respect to the replacement relationship $\simeq_r$ on sets of size $m$, {\small $\mathcal{M}^\prime:\mathbb{X}^n \to \mathbb{Y}$} satisfies {\small $(\log(1+(m/n)(e^\epsilon-1)), (m/n)\delta)$}-DP.
\end{lemma}
\section{FLAME Framework}
\noindent
In this section, we formalize the framework of \textbf{F}ederated \textbf{L}e\textbf{a}rning in the Shuffle \textbf{M}od\textbf{e}l, which we call FLAME. 

\subsubsection{Parties.}
\begin{figure}[t]
	\centering
	\includegraphics[width=0.465\textwidth,trim=2 62 2 66,clip]{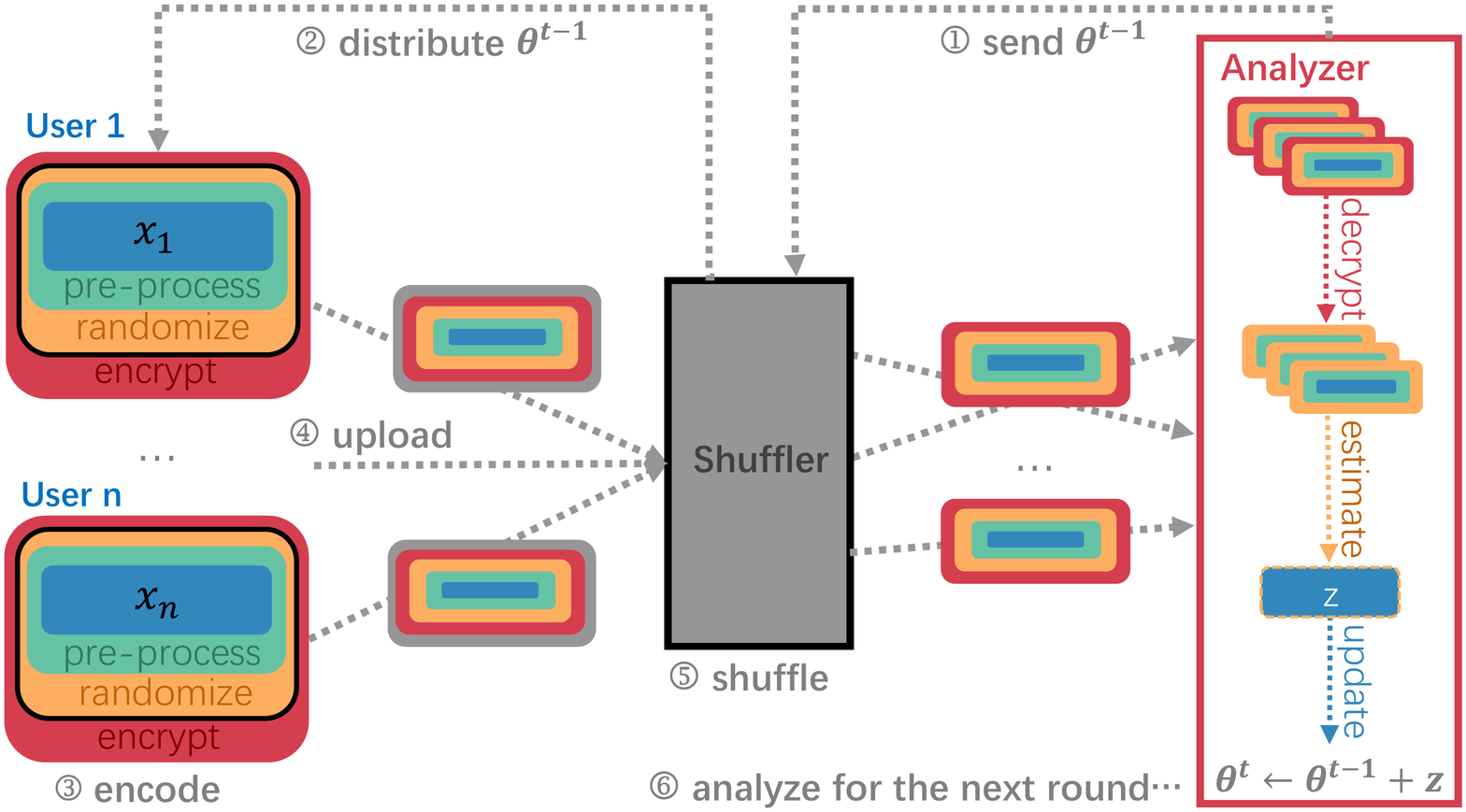}
	\caption{FLAME: Federated Learning in the Shuffle Model. 
	}\label{fig-overview}
\end{figure}

We present the FLAME architecture in Figure \ref{fig-overview} with three parties:
1) $n$ users, each of which owns a $d$-dimensional local update vector $x_i$ and runs a local randomizer $\mathcal{R}$ for the output  $y_i$.
2) The shuffler $\mathcal{S}$, the server who can perfectly shuffle received messages and send them to the analyzer.
3) The analyzer $\mathcal{A}$, the server that estimates the mean of shuffled messages into $z$ and updates the global model at the round $t$ by {\small  $\theta^t \leftarrow \theta^{t-1} + z$}.

\begin{table*}[t]
	\centering
	\begin{tabular}{|c|c|c|c|c|c|c|c|}
		\hline
		Parties                         & \multicolumn{3}{c|}{Shuffler $S$}                          & \multicolumn{3}{c|}{Analyzer $A$}                                       & Observer $\mathcal{O}$                       \\ \hline
		\multirow{2}{*}{Sensitive info} & \multicolumn{2}{c|}{gradient vector} & \multirow{2}{*}{ID} & \multicolumn{2}{c|}{gradient vector}              & \multirow{2}{*}{ID} & \multirow{2}{*}{query model}                 \\ \cline{2-3} \cline{5-6}
		& index                   & value      &                     & index                    & value                  &                     &                                              \\ \hline
		DP-FL                           & \multicolumn{3}{c|}{\multirow{2}{*}{N/A}}                  & \multicolumn{2}{c|}{$\surd$}                      & $\surd$             & \multirow{3}{*}{$(\epsilon_c, \delta_c)$-DP} \\ \cline{1-1} \cline{5-6}
		LDP-FL                          & \multicolumn{3}{c|}{}   & \multicolumn{2}{c|}{$(\epsilon_c, \delta_c)$-LDP} & $\surd$             &                                              \\ \cline{1-7}
		FLAME                           & $\times$                & $\times$   & $\surd$             & \multicolumn{2}{c|}{$(\epsilon_c, \delta_c)$-DP}  & $\times$            &                                              \\ \hline
	\end{tabular}\\
	$\surd$: trusted,
	$\times$: untrusted.
	\caption{Separation of Trust between Shuffler and Analyzer.}\label{tab-privacy}
\end{table*}

\subsubsection{Trust Boundaries.} 
We first clarify the trust boundary in FLAME.
We denote the observer as $\mathcal{O}$ which could be any curious party who can observe the global model parameters.
In the curator model (DP-FL), the trust boundary lies between $\mathcal{A}$ and $\mathcal{O}$.
In the local model (LDP-FL), the trust boundary lies between each individual user and the rest parties.
By introducing a shuffler $\mathcal{S}$, FLAME avoids placing full trust in any single party as DP-FL and meanwhile be able to achieve better utility than LDP-FL.

\subsubsection{Trust Separation.} 
Further, we clarify what are the private information and who can touch them in FLAME.
We design a fine-grained scheme of trust separation for FLAME and compare with DP-FL and LDP-FL in Table \ref{tab-privacy}.
Specifically, we separate the information of each local update into: the indexes, corresponding values and the user identity (i.e., ID in Figure \ref{fig-overview}).
It should be noted that the indexes could be sensitive when they are selected and sent to the shuffler $\mathcal{S}$ in a value-dependent manner.
Thus, our privacy goal in FLAME is to select indexes in a data-oblivious way and the true values of gradients are  invisible to $\mathcal{S}$.
But $\mathcal{S}$ should know users' implicit identities in order to distribute global model parameters and receive local messages.
The shuffled messages from $\mathcal{S}$ do not reveal the user identity and satisfy $(\epsilon_c, \delta_c)$-DP against $\mathcal{A}$.
For $\mathcal{O}$, $(\epsilon_c, \delta_c)$-DP holds by the post-processing property \cite{dwork2014the(book)}.
In LDP-FL, a $(\epsilon_c, \delta_c)$-LDP $\mathcal{R}$ is required by each user for this goal.

\subsubsection{FLAME framework.}
We present our framework in Algorithm \ref{alg-flame-overview} with three building processes: encoding $\mathcal{E}$, shuffling $\mathcal{S}$ and analyzing $\mathcal{A}$.
In Line  8, $C$ is the clipping threshold for the vector.
We  denotes the local privacy budget for each local vector by $\epsilon_l$.
$pk_a$ and $sk_a$ represent the public and secret key generated by the analyzer, respectively.
Different protocols below are designed by implementing functions \textsf{Randomize}($\cdot$) in Line 10 and \textsf{Shuffle}($\cdot$) in Line 15 with different strategies.
It should be noted that $\mathcal{R}_{\gamma, b}$ in Algorithm \ref{alg-krr-randomizer} can be applied in \textsf{Randomize}($\cdot$) as a basic randomizer, which accords to estimating with equation (\ref{eq-debias}) for line (\ref{line-analyze}).
Generic randomizers (e.g. Laplace Mechanism) can also be applied, which does not affect our corollaries later.

\subsubsection{Security Assumptions.}
We assume that the shuffler and the analyzer are not colluded (otherwise, FLAME is reduced to LDP-FL).
We also assume that the cryptographic primitives are safe and adversaries have computational difficulty to learn any information from the cipher text.

\subsubsection{Simple Protocol (SS-Simple).}\label{sec-v1}
We first propose SS-Simple $\mathcal{P} = \mathcal{A}\circ \mathcal{S}\circ \mathcal{R}^n$ for $d$-dimensional aggregation under the FLAME framework.
In a nutshell, we extend the one-dimensional protocol \cite{balle_privacy_2019} by conducting its randomization and aggregation for each dimension.
With the composition property in Lemma \ref{lemma-sequential-composition}, $\mathcal{R}$ should satisfy $\epsilon_{ld}$-LDP, where $\epsilon_{ld}=\epsilon_l/d$.
We instantiate \textsf{Randomize}($\cdot$) in Algorithm \ref{alg-flame-overview} for SS-Simple with $idx_i \leftarrow \{1, \cdots, d\}$ and $y_i \leftarrow \{\mathcal{R}_{\epsilon_{ld}}(x_{i, 1}), \cdots, \mathcal{R}_{\epsilon_{ld}}(x_{i,d})\}$.
The function \textsf{Shuffle}($\cdot$)  simply generates a permutation $\pi$ and outputs $m_{\pi(i)\in[n]}$.

Then we account for the central DP against the analyzer.
Take $\epsilon_{ld}$ to Lemma \ref{lemma-krr-bound} for $\mathcal{R}_{\gamma, b}$ or other numerical evaluations for generic $\mathcal{R}$, an amplified central privacy $(\epsilon_{cd}, \delta_{cd})$ can be derived.
With the composition in Lemma \ref{lemma-advanced-composition}, we can easily derive the vector-level composition in Theorem \ref{theo-simple-cd-d}.
Corollary \ref{coro-simple-l-c} distills the amplification from $\epsilon_l$ to $\epsilon_c$.
\begin{theorem}\label{theo-simple-cd-d}
	For any neighboring datasets $X \simeq_r X^\prime$ which differ in one user's $d$-dimensional local vector, $\mathcal{M}=\mathcal{S}\circ \mathcal{R}^n$ in SS-Simple satisfies $(\epsilon_c, \delta_c)$-DP, where:
	\begin{align}
	\epsilon_c &= d\epsilon_{cd} \land (\epsilon_{cd} \sqrt{2d \log(1/\delta_{cd})} + d\epsilon_{cd}(e^{\epsilon_{cd}} - 1) ),\\
	\delta_c &= \delta_{cd}(d+1).
	\end{align}
\end{theorem}
\begin{corollary}\label{coro-simple-l-c}
	For SS-Simple, with $\epsilon_l \le d\cdot \log(n/\log((d+1)/\delta_c))/2$, the amplified central privacy is $\epsilon_c = O( (1 \land \epsilon_l/d) e^{\epsilon_l/d} \log (d/\delta_c) \sqrt{d/n})$.
\end{corollary}

\subsubsection{Limitation of SS-Simple.}
Observing the Corollary \ref{coro-simple-l-c}, we find that the central DP level depends on the dimension $d$.
Intuitively, from the view of privacy, the amplification effect is diminished with a large $d$.
From the view of utility, randomizing the value $y_{i,j}$ for each dimension with a negligible privacy budget $\epsilon_{ld}=\epsilon_l/d$ incurs large noises.

\begin{algorithm}[t]
	\small
	\caption{FLAME: Encoding, Shuffling, Analyzing.}
	\begin{algorithmic}[1]\label{alg-flame-overview}
		\REQUIRE $\mathcal{A}, \mathcal{S}, \mathcal{E}, n, T, \epsilon_l, pk_a, sk_a$
		\ENSURE $\theta^T$
		\STATE Analyzer publishes $pk_a$
		\FOR {each iteration $t=1,\cdots, T$}
		\STATE Analyzer sends $\theta^{t-1}$ to Shuffler
		\STATE Shuffler distributes $\theta^{t-1}$ to a batch of $n$ users
		\FOR {each user $i\in[n]$ }
		\STATE $x_i\leftarrow \text{LocalUpdate}(\theta^{t-1})$
		\STATE {\color{black} $\rhd$ \texttt{Encoding $\mathcal{E}$ by each user}}
		\STATE $\bar{x}_i \leftarrow \text{Clip}(x_i, -C, C)$
		\STATE $\tilde{x}_i \leftarrow (\bar{x}_i+C)/(2C)$
		\STATE $\langle idx_i, y_i\rangle \leftarrow \textsf{Randomize}(\tilde{x}_i, \epsilon_l)$
		\STATE $c_i \leftarrow \text{Enc}_{pk_a}(y_i)$ \label{line-enc}
		\STATE user $i$ sends $m_i=\langle idx_i, c_i\rangle$ to Shuffler
		\ENDFOR
		\STATE {\color{black}$\rhd$  \texttt{Shuffling $\mathcal{S}$ by the Shuffler}}
		\STATE Shuffler sends $\textsf{Shuffle}(m_{i \in[n]})$ to Analyzer
		\STATE {\color{black} $\rhd$  \texttt{Analyzing $\mathcal{A}$ by the Analyzer}}
		\STATE decrypts values $y_{\pi(i) \in[n] } \leftarrow \text{Dec}_{sk_a}(c_{\pi(i) \in[n] })$ \label{line-dec} 
		\STATE estimates mean $\bar{z} \leftarrow \frac{1}{n} \sum_{i \in [n]} \langle idx_i, y_i \rangle$ \label{line-analyze}
		\STATE normalizes $z \leftarrow C\cdot(2\bar{z}-1) $
		\STATE updates model $\theta^t \leftarrow \theta^{t-1} + z $.
		\ENDFOR 
	\end{algorithmic}
\end{algorithm}

\section{Double Amplification (SS-Double)}
\subsubsection{Intuition.}
For strengthening the privacy amplification effect,
we propose an improved protocol SS-Double.
Instead of perturbing every dimension with a small budget $\epsilon_{ld}=\epsilon_l/d$, we only sample and perturb $k$ dimensions $k\ll d$.
As a result, each dimension can benefit from a larger privacy budget $\epsilon_{lk}=\epsilon_l/k$.
Furthermore, we notice that the privacy amplification can be further magnified by subsampling, which we call the \textit{double amplification}.
Intuitively, if the privacy amplification is strengthened, we could inject fewer noises under the same central privacy level.

\subsubsection{Challenge.}
However, the privacy amplification by subsampling may not be composable with shuffling for the multi-dimensional vector.
We first show how to compose the privacy amplification of shuffling and subsampling in one dimension with $\mathcal{R}_{\gamma, b}$.
Suppose $\mathcal{K}_r^{\beta}$ samples $n_s$ users from $n$ users with $\beta=n_s/n$.
The shuffler only receives encoded message from sampled users.
Applying Lemma \ref{lemma-krr-bound} and Lemma \ref{lemma-subsampling}, we derive Theorem \ref{theo-double-lk-cd-c}.
In addition, we should ensure $\delta_{cd} < 2\beta$ for a positive logarithm, which is reasonable to achieve since $\delta_{cd} \ll \frac{1}{dn}$ is negligible by standard.

\begin{theorem}\label{theo-double-lk-cd-c}
	With $\gamma=\frac{b}{e^{\epsilon_{ld}}+b-1}$ and $\delta_{cd} < 2\beta$, $\mathcal{M}=\mathcal{S} \circ \mathcal{R}^{n_s}_{\gamma, b} \circ \mathcal{K}_r^{\beta}$ satisfies $(\epsilon_{cd}, \delta_{cd})$-DP, where:
	\begin{equation}
	\epsilon_{cd} = \log(1+\beta(e^{\sqrt{\frac{14 \log(\frac{2\beta}{\delta_{cd}})(e^{\epsilon_{lk}}+b-1)}{n_s-1}}}-1)).
	\end{equation}
\end{theorem}

However, we cannot derive a similar theorem with $\mathcal{R}_{\gamma, b}$ in a multi-dimensional case. 
Intuitively, it is because the proof of privacy amplification by shuffling relies on bounded-size neighboring datasets, while subsampling may lead to two neighboring datasets with distinct size. 

\subsubsection{Dummy Padding.}
To solve the composition issue, we propose the method of \textit{dummy padding}: let the shuffler pad each dimension into the same size of $n_p$.
Denote the number of padding values for one dimension as $n_n=n_p-n_s$, and $n_n$ elements from the blanket distribution $\omega_\mathcal{R}$ are shuffled with all other messages received from users.
Thus, the SS-Double is $\mathcal{P}=\mathcal{A}\circ\mathcal{S}_p \circ \mathcal{R}^{n_p} \circ \mathcal{K}^\beta_r$,
where $\mathcal{S}_p$ consists of padding and shuffling by the shuffler and $\mathcal{K}_r^{\beta}$ denotes randomly subsampling by each user.
To instantiate SS-Double in our framework, the \textsf{Randomize}($\cdot$) works as follows.
First, it randomly samples $k$ indexes $idx_i \leftarrow \mathcal{K}_r^\beta(d)$.
For each index $j \in idx_i$, the perturbed value in $y_i$ is $y_{i,j}\leftarrow\mathcal{R}_{\epsilon_{ld}}(\tilde{x}_{i,j})$.
The procedures of \textsf{Shuffle}($\cdot$) are listed in Algorithm \ref{alg-double}.
\begin{algorithm}[t]
	\small
	\caption{\textsf{Shuffle}($\cdot$)=$\mathcal{S}_p$ for SS-Double}
	\begin{algorithmic}[1]\label{alg-double}
		\FOR {$j\in[d]$}
			\STATE $n_{s, j}\leftarrow \sum_{i=1}^n \mathbb{I}_{j\in idx_i}$
			\STATE $n_{n, j} \leftarrow n_p - n_{s, j}$
		\ENDFOR
		\STATE the number of dummy vectors $v \leftarrow \sum_{j=1}^d n_{n, j}/k$
		\FOR {$u\in [v]$}
			\STATE $S_{dummy} \leftarrow \{j|j\in[d], n_{n, j}\neq 0\}$
			\STATE $idx_{n+u} \leftarrow \{j|j\in S_{dummy}\}^k$
			\STATE $n_{n,j} \leftarrow n_{n, j} - 1$ for $j \in idx_{n+u}$
			\STATE $y_{n+u} \leftarrow \{y^*| y^*\leftarrow \omega_\mathcal{R}\}^k$
			\STATE $m_{n+u} \leftarrow \langle idx_{n+u}, \text{Enc}_{pk_a}(y_{n+u}) \rangle$
		\ENDFOR
		\STATE generates a permutation $\pi$ over $[nl+v]$
		\STATE shuffles and sends $\{m_{\pi(1)}, \cdots, m_{\pi(nl+v)}\}$ to analyzer
	\end{algorithmic}
\end{algorithm}

\subsubsection{Privacy and  Utility Analysis.}
We show the overall privacy amplification bound for $\mathcal{R}_{\gamma, b}$ in Theorem \ref{theo-double-lk-cd}.
We notice that a larger $n_p$ leads to a smaller $\epsilon_{cd}$, which implies a greater privacy amplification.
Correspondingly, a larger $n_p$ implies more noises injected as shown in Proposition \ref{prop-double-utility}.
\begin{theorem}\label{theo-double-lk-cd}
	With $\gamma=\frac{b}{e^{\epsilon_{lk}} +b-1}$ and $\delta_{cd}<2\beta$, $\mathcal{M}=\mathcal{S}_p \circ \mathcal{R}^{n_p}_{\gamma, b} \circ \mathcal{K}_r^{\beta}$ satisfies $(\epsilon_{cd}, \delta_{cd})$-DP, where:
	\begin{align}
	\epsilon_{ck} &= \sqrt{\frac{14\log(\frac{2\beta}{ \delta_{cd}})(e^{\epsilon_{lk}}+b-1)}{n_p-1}}, \\
	\epsilon_{cd} &= \log(1+\beta(e^{\epsilon_{ck}} - 1)).
	\end{align}
\end{theorem}
\begin{proposition}\label{prop-double-utility}
	The standard deviation of the estimated mean in each dimension from $\mathcal{P}=\mathcal{A}\circ \mathcal{S}_p \circ \mathcal{R}^{n_p}_{\gamma, b} \circ \mathcal{K}_r^{\beta}$ is $O(\frac{n_p^{1/6}\log^{1/3}(1/\delta_{ck}) }{n_s \epsilon_{ck}^{2/3}})$.
\end{proposition}

Since all dimension-level datasets are padded into the same size and $(\epsilon_{ck}, \delta_{ck})$-DP holds from the amplification in Theorem \ref{theo-double-lk-cd}, we show the vector-level DP composition in Theorem \ref{theo-double-cd-c}.
The sample rate is denoted as $\beta=k/d$.
We distill the amplification effect from $\epsilon_l$ to $\epsilon_c$ in Corollary \ref{coro-double-l-c}
\begin{theorem}\label{theo-double-cd-c}
	For any neighboring datasets $X \simeq_r X^\prime$ which differ in one user's local vector, the $d$-dimensional vector aggregation protocol $\mathcal{M} = \mathcal{S}_p \circ \mathcal{R}^{n_p} \circ \mathcal{K}_r^{\beta}$ in SS-Double satisfies $(\epsilon_c, \delta_c)$-DP, where:
	\begin{align*}
	\epsilon_c &= 2\beta d\epsilon_{cd}\land (\epsilon_{cd}\sqrt{4\beta d\log(1/\delta_{cd})}  + 2\beta d\epsilon_{cd}(e^{\epsilon_{cd}} -1)), \\
	\delta_c &= \delta_{cd}(2\beta d+1).
	\end{align*}
\end{theorem}

\begin{corollary}\label{coro-double-l-c}
	For SS-Double, with $\epsilon_l \le \beta d \log( n_p/\log((2\beta^2 d+\beta)/\delta_c) )/2$, the amplified central privacy is:
	\begin{equation}
	\epsilon_c = O( (1 \land \frac{\epsilon_l}{\beta d}) e^{\frac{\epsilon_l}{\beta d}} \beta^{1.5} \sqrt{\frac{d}{n_p}\log (\frac{\beta d}{\delta_c}) \log(\frac{\beta^2d}{\delta_{c}})}).
	\end{equation}
\end{corollary}

\subsubsection{Simulation of Privacy Amplification.}
\begin{figure}[t]
	\centering
	\includegraphics[width=0.5\textwidth,trim=0 0 0 0,clip]{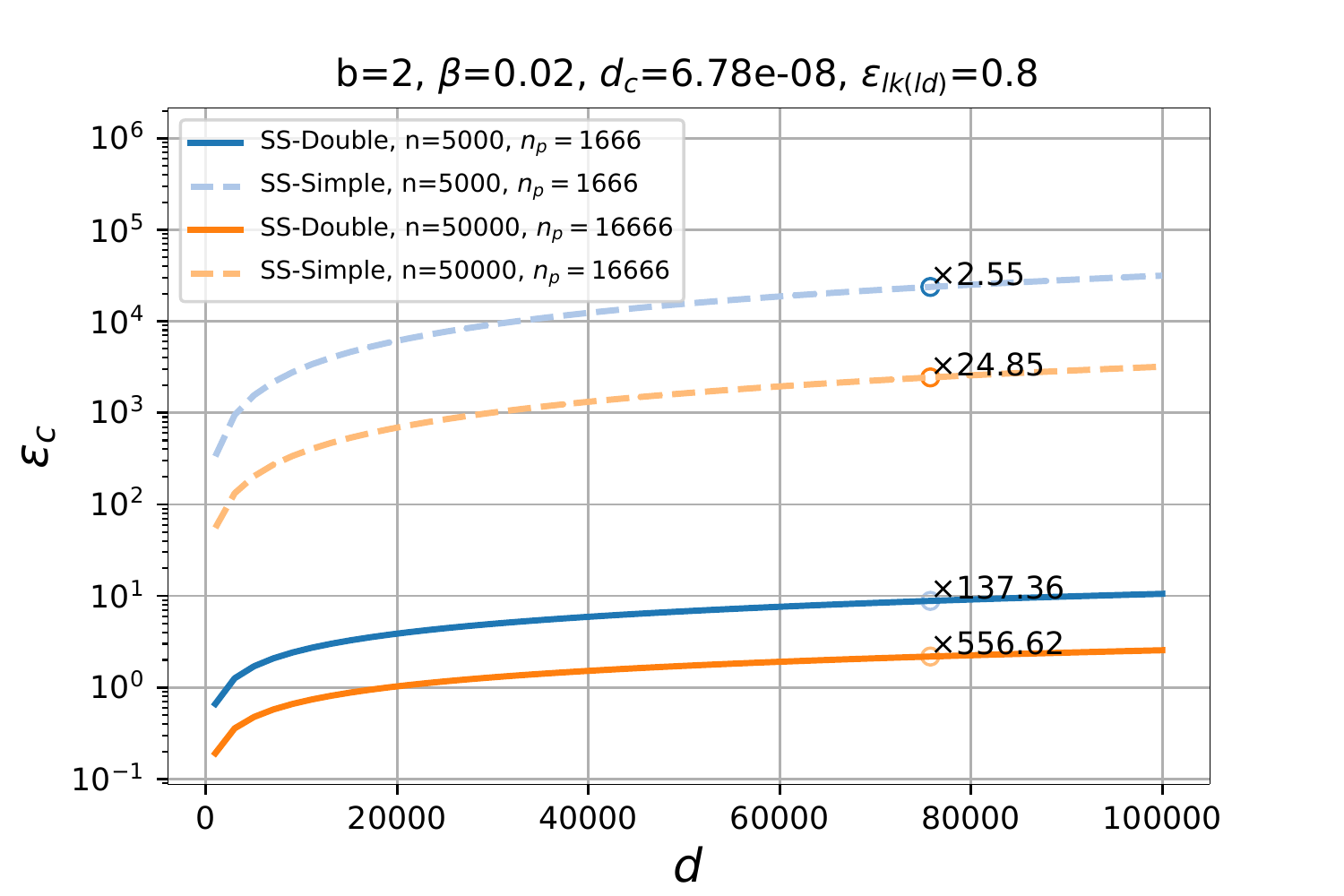}
	\caption{Privacy amplification effect comparison. 
	}\label{fig-krr-bound}
\end{figure}

To compare the privacy amplification effect of SS-Simple and SS-Double,
we visualize  the magnified privacy  w.r.t. $\mathcal{R}_{\gamma, b}$ in Figure \ref{fig-krr-bound}. 

\begin{enumerate}
	\item $\mathcal{S}\circ \mathcal{R}^n$ (SS-Simple): Lemma \ref{lemma-krr-bound} and Theorem \ref{theo-simple-cd-d}.
	\item $\mathcal{S}_p \circ \mathcal{R}^{n_p} \circ \mathcal{K}_r^{\beta}$ (SS-Double): Theorem \ref{theo-double-lk-cd} and Theorem \ref{theo-double-cd-c}.
\end{enumerate}
To align the valid condition in Lemma \ref{lemma-krr-bound}, we compare the above two cases given the same local privacy budget for \textit{each} perturbed dimension, denoted as $\epsilon_{ld}$ for SS-Simple and $\epsilon_{lk}$ for SS-Double. 
In this case, 
we present the amplification effect by the magnification ratio $\epsilon_l/\epsilon_c$ for $d=75755$.
It can be observed that the double amplification of SS-Double boosts improves the ratio from $\times2.55$ to $\times137.36$, which denotes a stronger privacy amplification effect.
It should be noted that the double amplification in Theorem \ref{theo-double-cd-c} also holds for other randomizers, which only differs with $\mathcal{R}_{\gamma, b}$ on the evaluation of $\epsilon_{ck}$. 
We show more numerical evaluations in experiments.

\section{Utility Boosting with Top-$k$ (SS-Topk)}
\subsubsection{Intuition.}
A problem of SS-Double protocol is that the random subsampling treats all dimensions equally and thus may discard ``important'' dimensions. 
For the high-dimensional case $(d>n)$, random sampling a small fraction $\beta$ of values from a vector slows down the convergence rate of training.
In light of the efficient gradient sparsification technique, we are motivated to adopt the magnitude-based selection \cite{aji2017sparse} for boosting the convergence rate.
However, selecting Top-$k$ indexes with greatest absolute magnitudes over the vector is data-dependent and thus compromises user privacy.
The challenge is how to preserve and qualify the \textit{index privacy} while maintaining the utility as far as possible.

\subsubsection{Index Privacy.}
According to our trust setting in  Table \ref{tab-privacy}, the prime adversary that threats index privacy is the shuffler $\mathcal{S}$ because only $\mathcal{S}$ knows which user sends which indexes (note that the perturbed value is encrypted).
Our goal is to bound the shuffler's success of predicting whether or not the index uploaded by a user ranks in the Top-$k$ elements of the local vector.
By random guessing, the adversary's success rate of predicting a dimension as Top-$k$ is $k/d$.
After observing the privatized selected indexes, the success rate cannot be enlarged by more than $\nu$ times.

Thus, we are motivated to qualify and preserve the index privacy with an anonymity-based metric in Definition \ref{def-top-index-privacy}.
We denote whether the magnitude of dimension $j\in[d]$ ranks in Top-$k$ ($\beta=k/d$) by $\mathbb{I}_j \in \{0,1\}$, and whether the index $j$ is selected by $\mathcal{K}_{\nu}^\beta(j)$.
The first inequality bounds the adversary's success rate with $\nu$ while the second inequality ensures the probability is no greater than 1.
Intuitively, the strongest index privacy stands when $\nu=1$, because observing $\mathcal{K}_{\nu}^\beta(j)$ does not increase the adversarial success rate.
\begin{definition}\label{def-top-index-privacy}
	A mechanism $\mathcal{K}_{\nu}^\beta$ provides $\nu$-index privacy for a $d$-dimensional vector, if and only if for any $j\in[d], \nu \ge 1$, we have:
	$\Pr[\mathbb{I}_j=1|\mathcal{K}_{\nu}^\beta(j)] \le \nu \cdot \Pr[\mathbb{I}_j=1]$ and   $\Pr[\mathbb{I}_j=0|\mathcal{K}_{\nu}^\beta(j)] \ge \frac{\Pr[\mathbb{I}_j=0]}{\nu}$. 
\end{definition}

To achieve $\nu$-index privacy, we need to control the probability of $\Pr[\mathbb{I}_j=1| \mathcal{K}_{\nu}^\beta(j)]$.
Given the prior knowledge $\Pr[\mathbb{I}_j=1]=\beta, \Pr[\mathbb{I}_j=0]=1-\beta$, if each user reports $lk$ dimensions to the shuffler of which only $k$ indexes are real Top-$k$, we have $\Pr[\mathbb{I}_j=1|\mathcal{K}_{\nu}^\beta(j)]=\frac{1}{l}$.
Put it into the definition, we have:
$\Pr[\mathbb{I}_j=1|\mathcal{K}_{\nu}^\beta(j)] \le \min(\nu \cdot \beta, \frac{\nu-1+\beta}{\nu})$.
Thus, $\mathcal{K}_{\nu}^\beta$ satisfies $\nu$-index privacy when we set $l\ge \max \{\frac{1}{\nu \beta}, \frac{\nu}{\nu-1+\beta}\}$.

\begin{algorithm}[t]
	\small
	\caption{\textsf{Randomize}($\cdot$)=$\mathcal{R} \circ \mathcal{K}_{\nu}^\beta$ for SS-Topk}
	\begin{algorithmic}[1]\label{alg-topk}
		\STATE Choose a valid $l \in [\max \{\frac{1}{\nu \beta}, \frac{\nu}{\nu-1+\beta}\}, \lceil 1/\beta \rceil]$
		\STATE Top-$k$ index set $S_{top} \leftarrow \{j|j\in \text{Top}(|\tilde{x}_i|)\}^k $
		\STATE Non-Top index set $S_{non} \leftarrow \{j|j\in [d]\backslash S_{top}\}^{k(l-1)}$
		\FOR {each index $j\in S_{top} \cup S_{non}$}
		\STATE
		$
		y_{i,j} \leftarrow
		\begin{cases}
		\mathcal{R}_{\epsilon_{lk}}(\tilde{x}_{i, j})
		, & j \in S_{top} \\
		\omega_\mathcal{R}
		, & j \in S_{non}
		\end{cases}
		$
		\ENDFOR
		\STATE generate a permutation $\pi_r$ over $[lk]$
		\STATE the index list $idx_i \leftarrow \{d_{\pi_r(1)}, \cdots, d_{\pi_r(lk)}\}$
		\STATE the perturbed value list $y_i \leftarrow \{y_{\pi_r(1)}, \cdots, y_{\pi_r(lk)}\}$
		\STATE returns $\langle idx_i, y_i \rangle$
	\end{algorithmic}
\end{algorithm}

\subsubsection{SS-Topk Protocol.}
Denote the SS-Topk protocol as $\mathcal{P}=\mathcal{A} \circ \mathcal{S}_p \circ \mathcal{R}^{n_p} \circ  \mathcal{K}_{\nu}^\beta$.
Each user locally runs $\mathcal{R} \circ  \mathcal{K}_{\nu}^\beta$.
The shuffler executes $ \mathcal{S}_p$.
The analyzer runs the aggregation $\mathcal{A}$.
Compared with SS-Double, SS-Topk has the same shuffling and analyzing procedure but differs in the dimension selection $\mathcal{K}_{\nu}^\beta$ in \textsf{Randomize}($\cdot$) which is shown in Algorithm \ref{alg-topk}.

By applying $\mathcal{K}_{\nu}^\beta$ on the processed vector $\tilde{x}_i$,
$k$ Top-$k$ indexes are sampled as the set $S_{top}$ while $k(l-1)$ non-top dimensions are randomly sampled from the rest as $S_{non}$.
As $k$ true value are perturbed, the privacy budget split for each dimension is $\epsilon_{lk}=\epsilon_l/k$.
Then each value of real Top-$k$ dimensions is perturbed as $y_{i,j} \leftarrow \mathcal{R}_{\epsilon_{lk}}(\tilde{x}_{i,j})$.
Each non-top dimension is padded with a value drawn from the blanket distribution $\omega_\mathcal{R}$. 
Then $kl$ dimensions are permuted into the list of $idx_i$ and $y_i$, which will be encrypted into $l$ messages by Line (\ref{line-enc}) in Algorithm \ref{alg-flame-overview}.


\begin{figure}[h]
	\centering
	\includegraphics[width=0.4\textwidth,trim=0 0 0 0,clip]{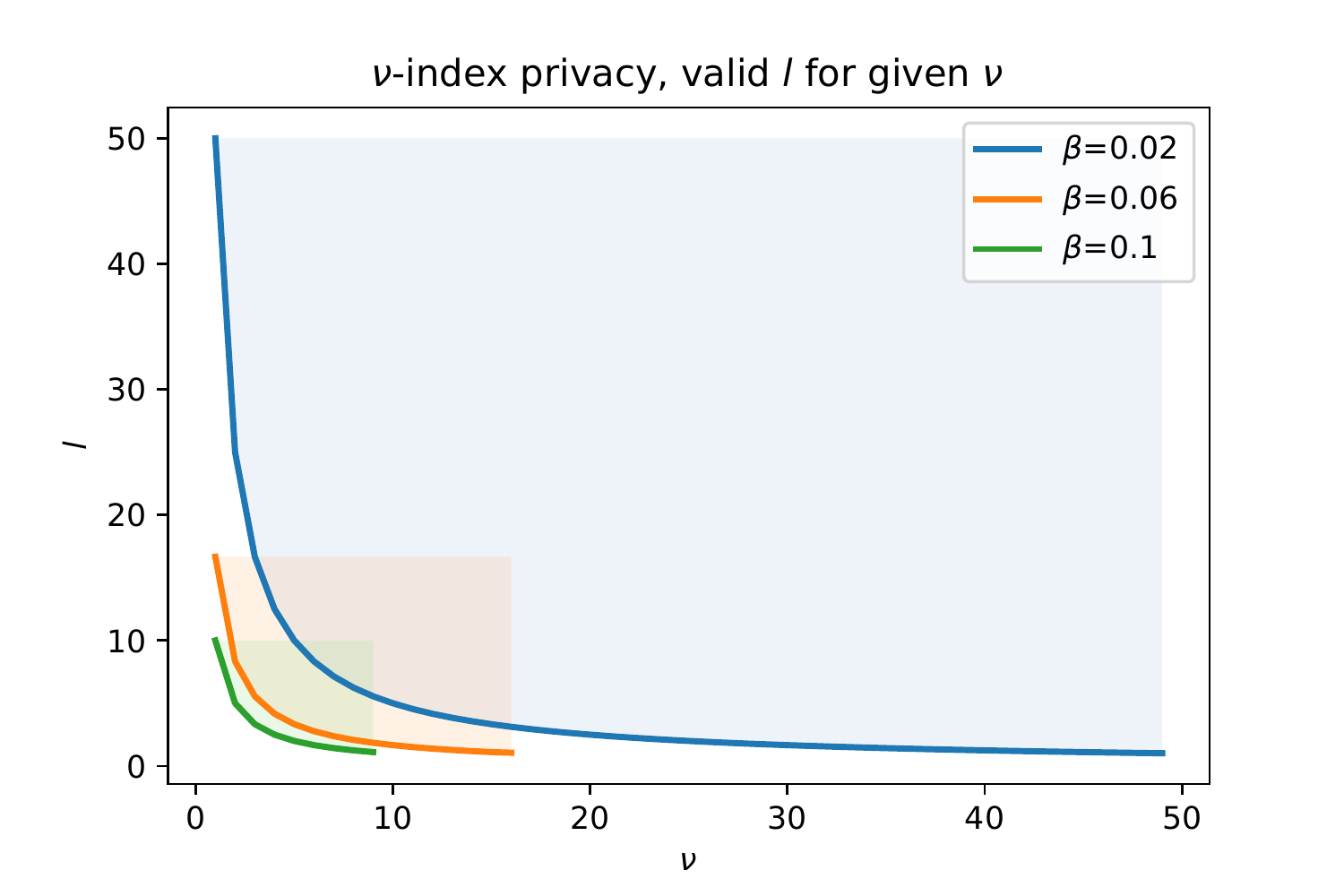}
	\caption{Smoothing between $\nu$ and $l$. 
	}\label{fig-index}
\end{figure}

\subsubsection{Privacy Analysis}
First, we show the relationship of $\nu$ and $l$ in Proposition \ref{prop-topk-smooth}.
The valid $l$ for a given $\nu$ is illustrated in Figure \ref{fig-index}.
The strongest index privacy $\nu=1$ is achieved when $l=\lceil 1/\beta \rceil$.
For SS-Double, $\nu$ naturally holds with the random sampling mechanism $\mathcal{K}_r^\beta$.
To the other extreme case of no index privacy, SS-Topk still provides a strong privacy guarantee since the shuffler knows nothing except for Top-$k$ indexes.
An state-of-the-art work \cite{zhu2019deep} has shown the privacy attack's availability is significantly impaired even knowing both Top-$k$ indexes as well as their values.
\begin{proposition}\label{prop-topk-smooth}
	The range of $\nu$-index privacy is $1 \le \nu \le \frac{1}{\beta}$, where the strongest index privacy $\nu=1$ is achieved when $l=\lceil \frac{1}{\beta} \rceil$ and no index privacy is achieved when $l=1$.
\end{proposition}

Then, we clarify the compatibility of $\nu$-index privacy against the shuffler and $(\epsilon_c, \delta_c)$-DP against the analyzer.
With $\mathcal{S}_p$ in Algorithm \ref{alg-double}, the analyzer only gets padded results for each dimension with the same size $n_p$.
Thus, the $\nu$-index privacy against the shuffler \textbf{does not} affects the amplified privacy $(\epsilon_c, \delta_c)$-DP against the analyzer.
Given $n_p$, SS-Topk shares the same double amplification effect of SS-Double.
Lastly, we discuss the trade-off between the index privacy and the communication costs as well as the utility.
The bandwidth of each user depends on $O(lk)$.
As Proposition \ref{prop-double-utility} implies, the estimation utility depends on the dummy padding size $n_p$.
Thus, given $n_p, n$ and $\beta$, $\nu$ \textbf{does not} affect the accuracy of the mean estimation because the number of dummy values is fixed.
We show the strongest index privacy under the given parameters in Theorem \ref{theo-topk-mp-l}.

\begin{theorem}\label{theo-topk-mp-l}
	Given a protocol with $\mathcal{K}_{\nu}^\beta, n_p$, the strongest index privacy it allows for each user is $\nu=\max\{1, \frac{1}{  \lfloor \frac{n_p}{n\beta} \rfloor \cdot \beta }\}$.
\end{theorem}
\section{Evaluations}
\noindent
We evaluate the learning performance on MNIST dataset and logistic regression model with $d=7850, n=1000$.
Baselines include non-private Federated Averaging (NP-FL) \cite{mcmahan2016Federated}, DP-FL \cite{abadi2016deep} with Gaussian Mechanism in which we double the sensitivity for the comparison under bounded DP definition, LDP-FL with Gaussian Mechanism \cite{Dwork2006Our} for $(\epsilon, \delta)$-LDP and Laplace Mechanism for $\epsilon$-LDP and our three protocols of SS-Simple, SS-Double, SS-Topk \footnote[1]{Code is available at: https://github.com/Rachelxuan11/FLAME}.

\begin{figure}[t]
	\begin{minipage}[t]{0.5\linewidth}
		\centering
		\includegraphics[width=\textwidth,trim=5 10 10 5,clip]{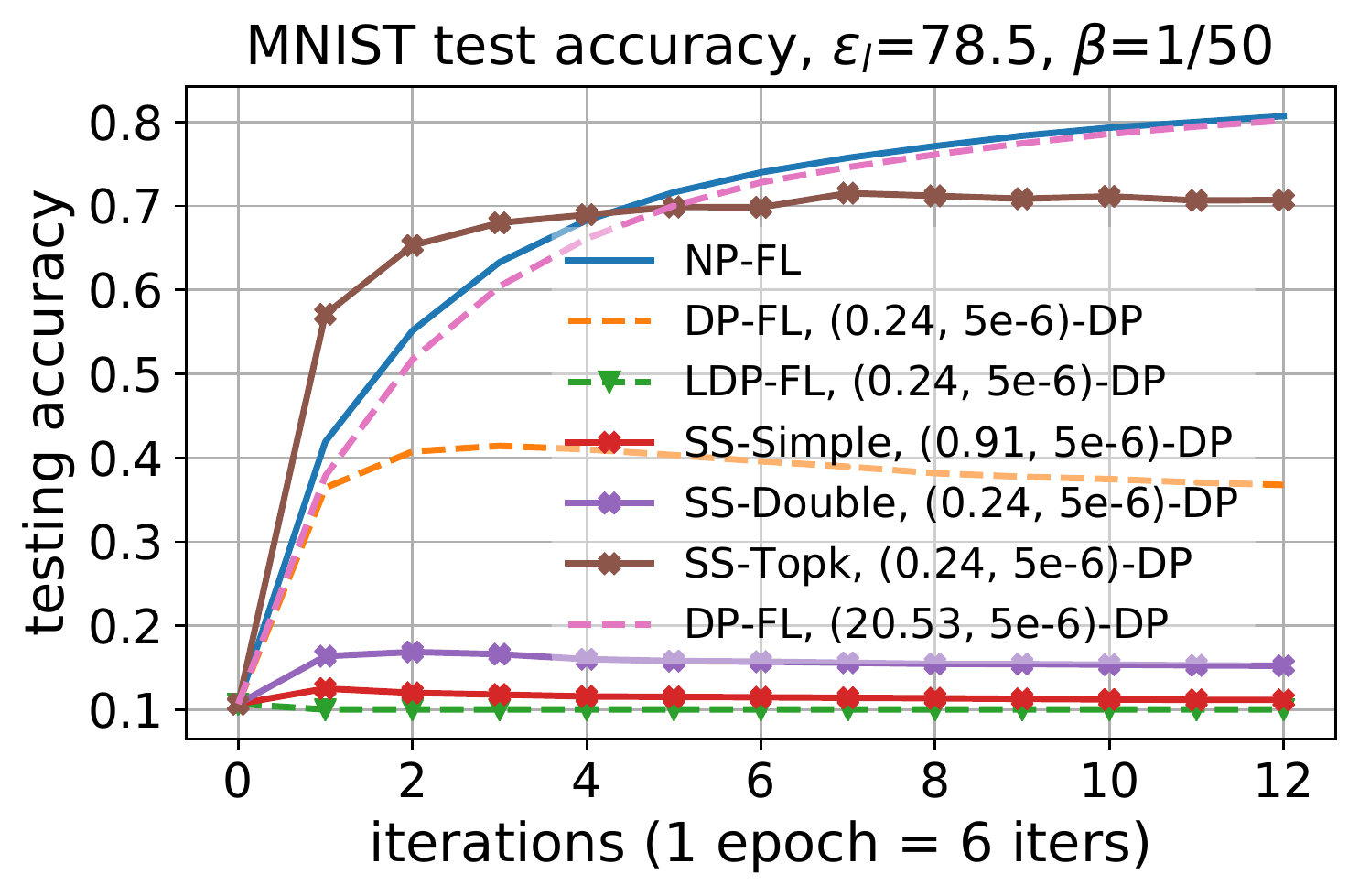}
		\caption{Utility-Privacy.}\label{fig-baseline}
	\end{minipage}%
	\begin{minipage}[t]{0.5\linewidth}
		\centering
		\includegraphics[width=\textwidth,trim=5 10 10 5,clip]{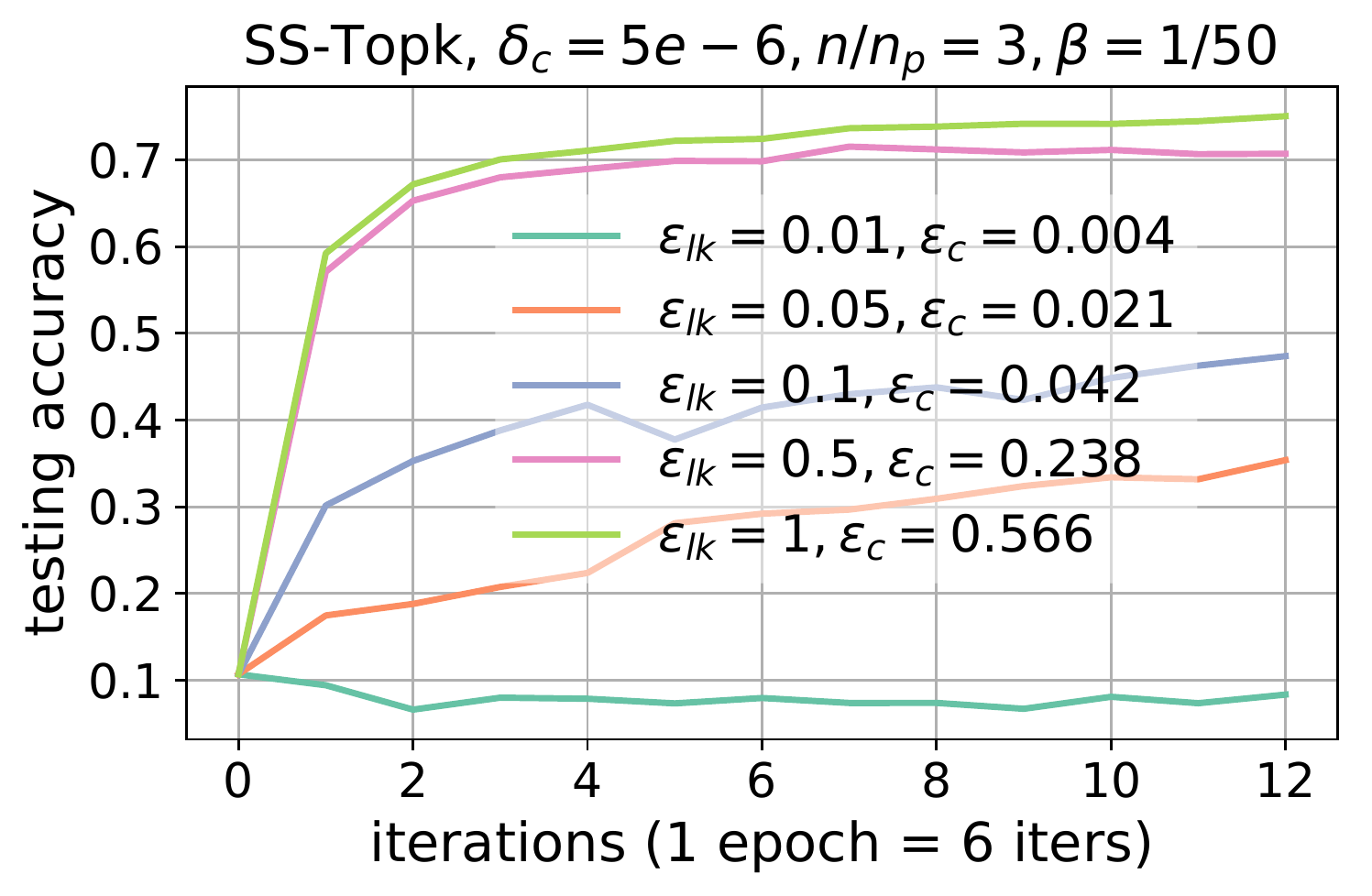}
		\caption{Impact of $\epsilon_{lk}$.}\label{fig-eps}
	\end{minipage}%
\end{figure}

\begin{figure}[t]
	\begin{minipage}[t]{0.5\linewidth}
		\centering
		\includegraphics[width=\textwidth,trim=5 10 10 5,clip]{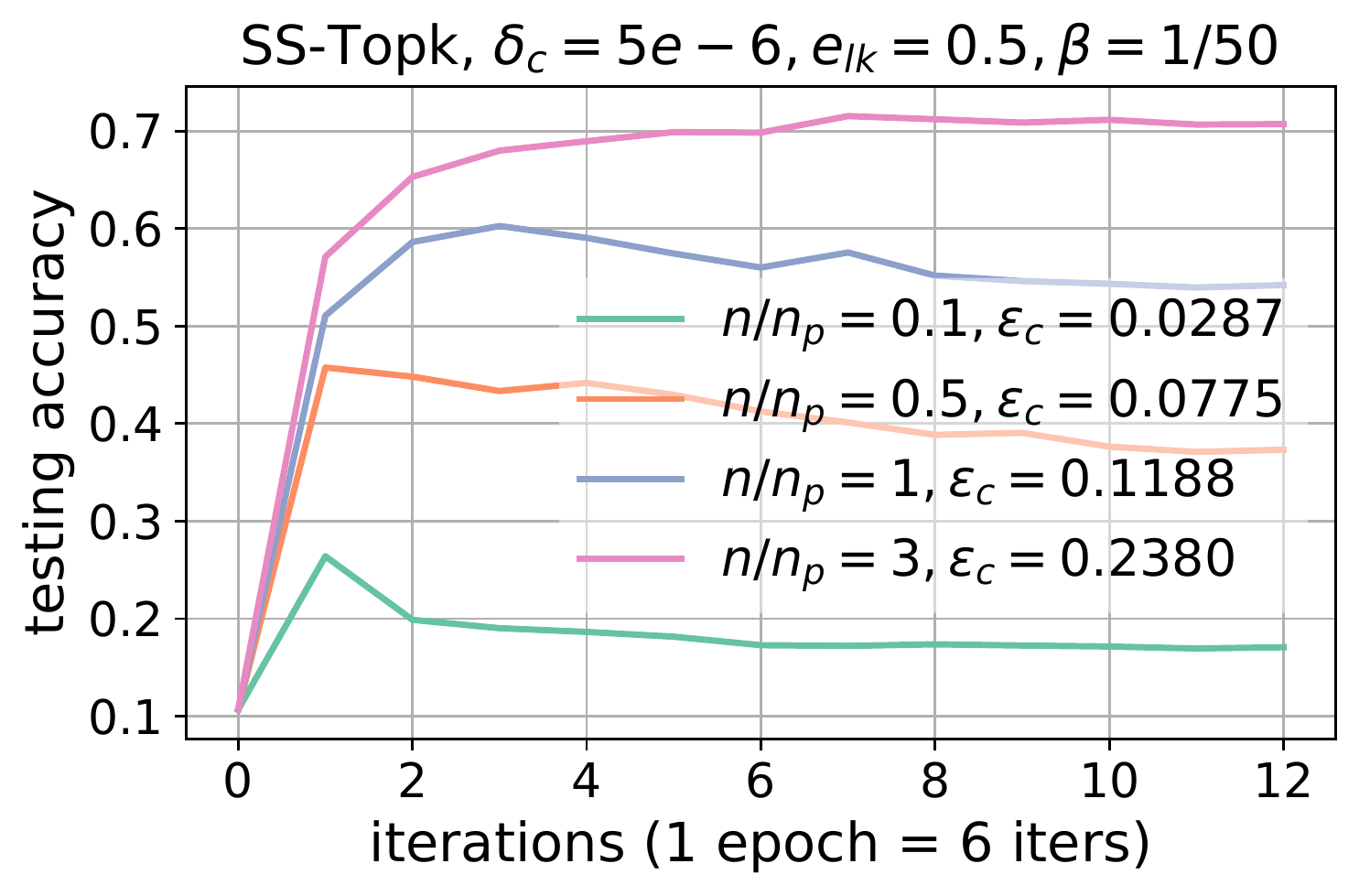}
		\caption{Impact of $n/n_p$.}\label{fig-mp}
	\end{minipage}%
	\begin{minipage}[t]{0.5\linewidth}
		\centering
		\includegraphics[width=\textwidth,trim=5 10 10 5,clip]{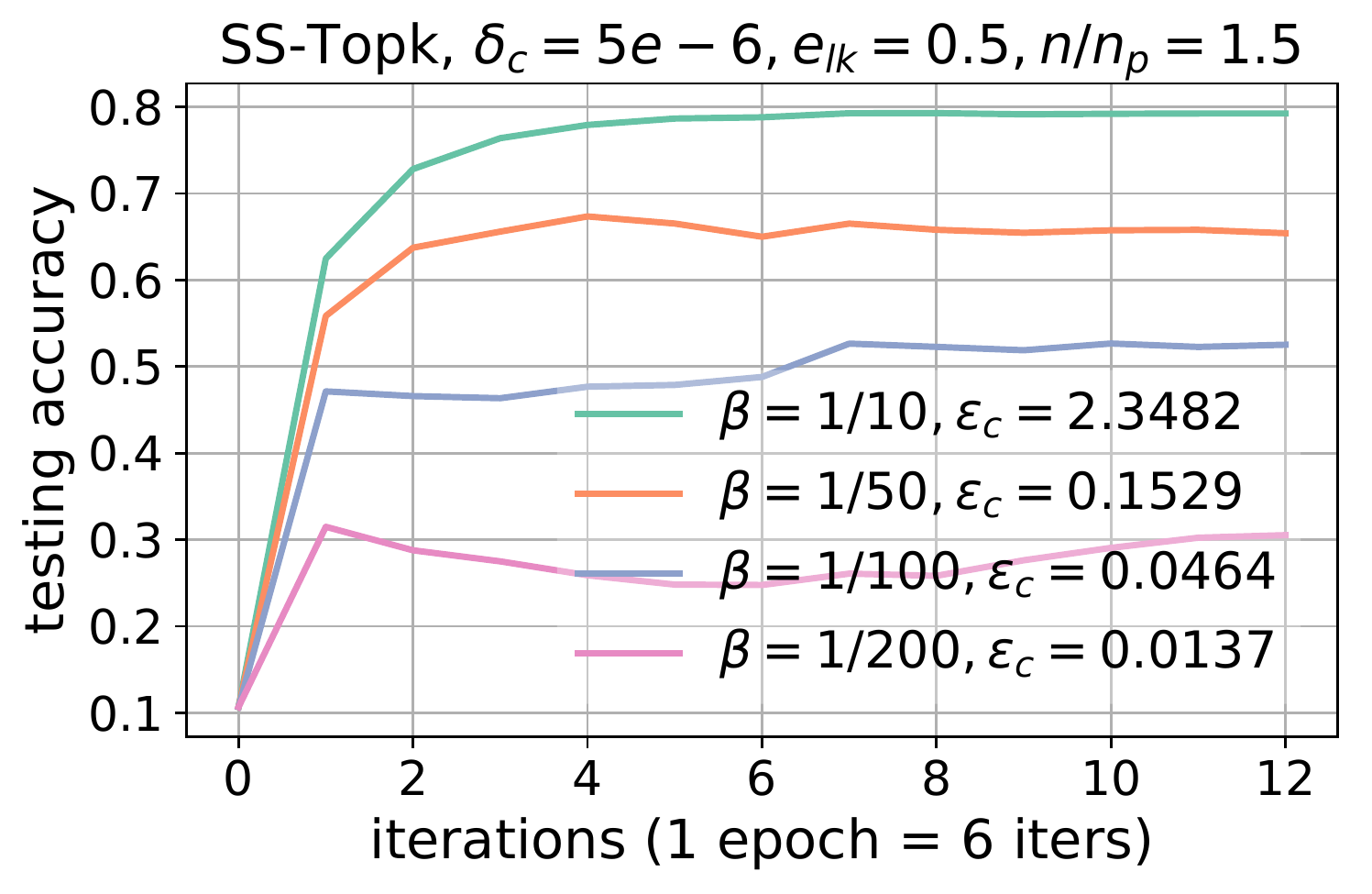}
		\caption{Impact of $\beta=k/d$.}\label{fig-rate}
	\end{minipage}%
\end{figure}

\begin{figure}[t]
	\begin{minipage}[t]{0.5\linewidth}
		\centering
		\includegraphics[width=\textwidth,trim=5 0 0 5,clip]{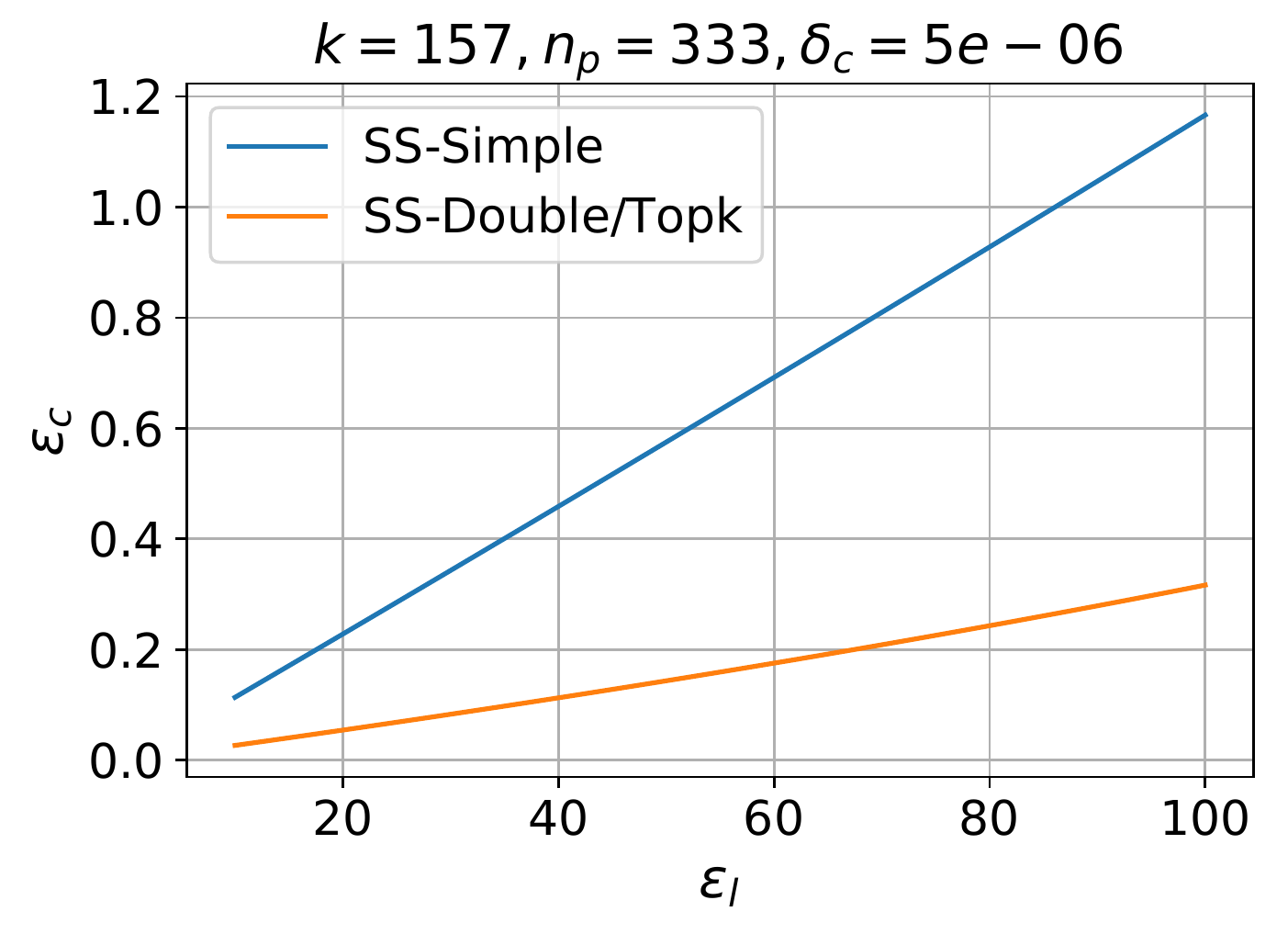}
		\caption{Under various $\epsilon_l$.}\label{fig-el-lap-ben}
	\end{minipage}%
	\begin{minipage}[t]{0.5\linewidth}
		\centering
		\includegraphics[width=\textwidth,trim=10 0 6 6,clip]{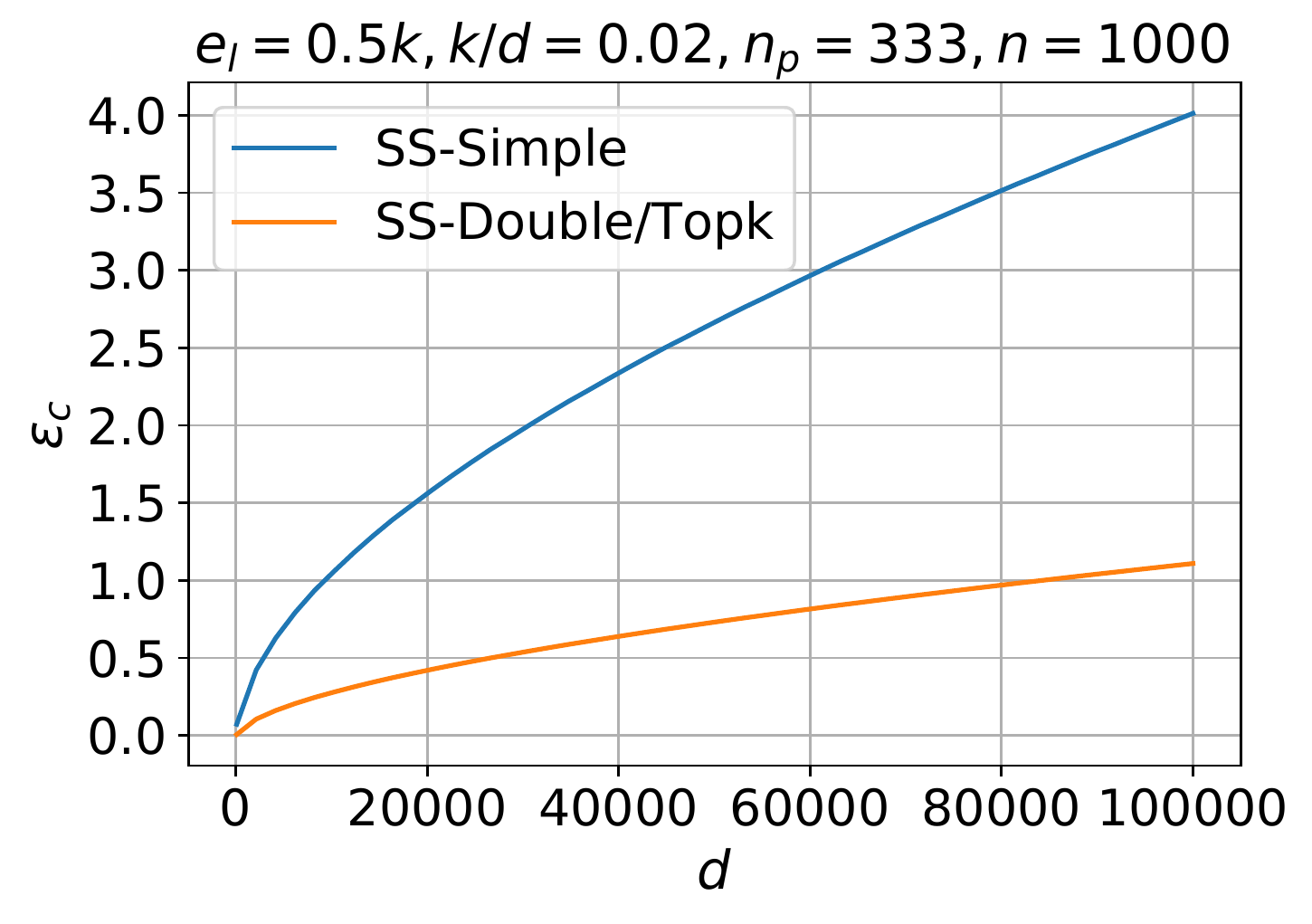}
		\caption{Under various $d$.}\label{fig-d-lap-ben}
	\end{minipage}%
\end{figure}

\subsubsection{Comparison of Our Protocols.}
For SS-Simple, SS-Double, SS-Topk, we apply the Laplace Mechanism as the basic randomizer $\mathcal{R}$ for each dimension.
Given $\epsilon_l=78.5$, the spilt privacy budget of each dimension is $\epsilon_{ld}=0.01$ for SS-Simple and $\epsilon_{lk}=0.5$ for SS-Double and SS-Topk.
The double amplification effect boosts the amplified privacy against $\mathcal{A}$ for one epoch from $(0.91, 5\times 10^{-6})$-DP (SS-Simple) to $(0.24, 5\times 10^{-6})$-DP (SS-Double/Topk).
Compared with SS-Simple, SS-Double improves the testing accuracy by 4.07\% under a stronger central privacy.
Compared with SS-Double, SS-Topk significantly boosts the utility by 55.5\% under the same central privacy.
With Theorem \ref{theo-topk-mp-l} and $n/n_p=3$ in Figure \ref{fig-baseline}, the maximum index privacy that the protocol allows for each user is $\nu=3.125$ with $l=16$.

\subsubsection{Comparison with DP/LDP-FL.}
It is obvious that even the baseline SS-Simple performs better than LDP-FL.
Then we observe that under the same central privacy $(0.24, 5\times 10^{-6})$-DP, SS-Topk even achieves a dramatic higher accuracy than DP-FL by 33.94\%.
This is a \textbf{key observation} of our work, because traditional works for the one-dimensional task claim that the shuffle model only stands in the middle-ground between LDP and DP.

We analyze the reasons as follows:
1) The effect of Top-$k$: as SS-Double cannot exceed the performance of DP-FL $(0.24, 5\times 10^{-6})$-DP but SS-Topk can, it is obvious that Top-$k$ boosts the utility.
2) The effect of the double amplification:
If we only counts the amplification from shuffling for SS-Topk without the amplification from subsampling, we have $\epsilon_c=20.53, \delta_c=5\times 10^{-6}$.
Hence, we introduce another baseline of DP-FL $(20.53, 5\times 10^{-6})$-DP.
We observe that this line approaches the non-private version and has higher testing accuracy than SS-Topk\footnote[2]{SS-Topk satisfies $(20.53, 5\times 10^{-6})$-DP when ignoring subsampling amplification effect.}.
In other words, SS-Topk cannot perform better that DP-FL if only amplification of shuffling is counted.
Thus, we validate the effect of proposed double sampling for such nontrivial utility boosting.
Thus, we conclude that \textbf{both} double amplification and Top-$k$ boosting are necessary to performs better than DP-FL.

\subsubsection{Comparison under Variant Parameters.}
We then evaluate the impacts of other hyper-parameters of $n_p, \beta$ and privacy budget $\epsilon_l$ in Figures \ref{fig-eps}, \ref{fig-mp} and \ref{fig-rate}.
1) It is obvious in Figure \ref{fig-eps} that a larger local privacy budget for each dimension leads to higher testing accuracy.
2) In Figure \ref{fig-mp}, the higher ratio of $n/n_p$ implies less additional noises injected by \textit{dummy padding}. 
This validates our utility analysis in Proposition \ref{prop-double-utility}.
Thus, we can conclude that $n_p$ is the key knob to tune the privacy and utility trade-off.
3) In Figure \ref{fig-rate}, we can observe that larger $\beta$ implies better utility.
With $\beta=1/10$ in Figure \ref{fig-baseline}, we can achieve only 1.48\% loss compared with NP-FL for 2 epochs of $(2.348, 5e^{-6})$-DP per epoch.

\subsubsection{Privacy Amplification.}
We illustrate the overall amplification in previous evaluations with the Bennett inequality for the Laplace Mechanism in Figure \ref{fig-el-lap-ben} and Figure \ref{fig-d-lap-ben}.
This validates our Theorems \ref{theo-simple-cd-d}, \ref{theo-double-cd-c} and Corollaries \ref{coro-simple-l-c}, \ref{coro-double-l-c} are generic for any local randomizer, as long as the amplification bound can be derived from a close-form solution or numerical evaluations \cite{balle_privacy_2019}.
\section{Conclusion}
\noindent
To conclude, we propose the first differentially private federated learning framework FLAME in the shuffle model for better utility without relying on any trusted server.
Our privacy amplification effect and private Top-$k$ selection mechanism significantly boosts the testing accuracy under the high-dimensional setting.
\section{Related Works}
\subsection{Private-preserving Federated Learning}
In a single iteration in the cardinal federated learning framework \cite{mcmahan2016Federated}, the analyzer first distributes a global model, then a batch of $n$ users updates local models in parallel.
Let $x_i\in \mathbb{R}^d$ denote a local update.
The analyzer averages $x_i,\cdots, x_n$ to update the global model.
$
\theta^t \leftarrow \theta^{t-1} + \frac{1}{n} \sum_{i=1}^{n} x_i.
$

\subsubsection{FL in the Curator Model (DP-FL).}
Distributing global parameters has been a critical interface of privacy leakages \cite{zhu2019deep,shokri2017membership}.
The curator DP model's definition ensures whether a specific user participants in the federated training is not revealed by distributed model parameters.

Gaussian mechanism \cite{abadi2016deep,mcmahan_learning_2018} is the cardinal mechanism to distort the averaged gradient for $(\epsilon_c, \delta_c)$-DP.
Given the maximum $L_2$ norm $C$, each global vector $x_i$ is first clipped to $\bar{x}_i = x_i \cdot \min \{1, C/||x||_2 \} $ for bounding the privacy.
Then i.i.d. sampled noises are injected to each dimension:
$\theta^t \leftarrow \theta^{t-1} + \frac{1}{n} ( \sum_{i=1}^{n} \bar{x}_i + \mathcal{N}(0; \sigma^2 \mathbb{I}_d) )$, where $\sigma = \frac{\bigtriangleup_2}{\epsilon_c} \sqrt{2\ln 1.25/\delta_c}$. 
With neighboring datasets $X^\prime\simeq_r X$ by replacing any single user, we have $\bigtriangleup_2=2 C$.
However, the analyzer in the curator DP model is assumed to be honest and obey the instructions of private mechanisms.
Intuitively, a distributed way \cite{agarwal2018cpsgd} is to splitting the required noise $\mathcal{N}(0, \sigma^2\mathbb{I}_d)$ to $n$ shares for each user.
Then the aggregated noise can provide $(\epsilon_c, \delta_c)$-DP.
But such distributed noise only provides weak local differential privacy against the untrusted analyzer unless a heavy cryptography protocol can be applied or the $\epsilon_c$ is extremely small.

\subsubsection{FL in the Local Model (LDP-FL).}
For a formal privacy guarantee against the analyzer, it is natural to perturb each user's local vector with $\epsilon_l$-LDP.
However, LDP is notorious for the low utility, especially for the multi-dimensional vector, because its mean estimation error is bounded by $O(\frac{\sqrt{d\log d}}{\epsilon_l \sqrt{n}})$. 
This dimension curse on $d$ implies that we need to trade a better utility with either loose privacy guarantee or unacceptable large user sample size.
Besides, if stricter privacy $(\epsilon_c, \delta_c)$-DP is required for model publishing $(\epsilon_c<\epsilon_l)$,
additional noises are required for the average  \cite{bhowmick2018protection}, which will further ruin the utility.

\subsubsection{Shuffle Model.}
The shuffle model is proposed in PROCHLO \cite{bittau_prochlo_2017}, which is a general framework of encode-shuffle-analyze (or ESA).
It generalizes the local and central models by setting a non-collude party to execute random shuffling of encrypted encodings before sending the analyzer.
The theoretical amplification effect is firstly proved in \cite{erlingsson_amplification_2018} for interactive locally differential private protocol under a limited regime of the privacy budget.

To extending the results for more general regime, \cite{cheu_distributed_2019} specialize the amplification to the one-bit randomized response protocol.
They show that the shuffle model cannot provide a more accurate result than the central model, but can achieve more accurate real summation estimation than the local model.
To improve a better privacy amplification bound, \cite{balle_privacy_2019} proposes an optimal \textit{single-message} protocol with theoretical proof framework named as the ``Privacy Blanket''.
They also provide a new lower bound for the accuracy for their real-summation protocol.
In stead of transmitting with only one message for each user, \textit{multi-message} protocols are discussed \cite{cheu_distributed_2019,ghazi_scalable_2019,balle_differentially_2019,ghazi2020pure}.
When more messages sent to the shuffler, the analyzer can obtain more accurate estimation results than \textit{single-message} protocols.
Our work is based on the \textit{single-message} protocol, it is promising to extending it to the \textit{multi-message} setting for better accuracy with more communication costs.
As the building block, the shuffle model protocol for one-dimensional summation can be extended to other tasks, such as the histogram estimation \cite{ghazi2020pure, balcer2019separating,wang2019murs}.

Contemporaneous works \cite{erlingssonEncodeShuffleAnalyze2020, ghazi_scalable_2019} discuss FL in the general ESA framework but have different trust assumptions from this work.
Recently, \cite{girgisShuffledModelFederated2021} propose a communication efficient federated learning framework. They utilize the amplification effect from subsampling and shuffling, which is compatible but orthogonal to our methods since they sample clients and local records. Also, the trust model against the shuffler is different from ours.
\section{Ethics Statement}
\noindent
In this paper, the authors introduce FLAME, a framework in the shuffle model for differentially private federated learning without any trusted parties.
	Federated Learning can be applied to a wide range of applications, including the next word predicting, fraud detection model, training an epidemic prediction model, and many more \cite{yang2019Federated}.
	
	There will be essential impacts resulting from federated learning in general. 
	Here we focus on the effects of using FLAME to preserve privacy for the cross-device setting, where a global model is trained over massive personal data that are generated on small gadgets or mobile devices.
	Our research could be used to provide better learning performance under a strong privacy guarantee in these applications.
	Thus, it will promote the information flow across millions of mobile devices while strictly maintaining users' human rights for data privacy (GDPR:https://gdpr-info.eu/).
	
	The potential risks of differentially private federated learning over mobile devices have typically received less attention.
	These include: 
	\begin{itemize}
	\item There is a lack of concrete law or regulation for the privacy budget over iterative rounds. It should be transparent for users knowing their privacy budget setting to feel safe and be safe.
	\item If differential privacy is applied for local updates, this could increase the difficulty of detecting adversarial attacks from the user end, leading to a biased global model.
	\end{itemize}
	
	Besides, we see opportunities for research related to shuffle model FLAME to beneficial purposes, such as providing better product services for users and better monitoring the financial crisis and a public health emergency (e.g. COVID-19).
	To mitigate the risks associated with using differentially private federated learning, we encourage research to understand the impacts of robustness when applying models to scenarios where the prediction accuracy is highly sensitive.

\section{Acknowledgements}
\noindent
This work is supported by 
the National Natural Science Foundation of China (No. 62072460, No. 61532021, 61772537, 61772536, 61702522), JSPS KAKENHI Grant No. 17H06099, 18H04093, 19K20269, JST/NSF Joint Research SICORP 20-201031504 and CCF-Tencent Open Fund WeBank Special Fund.


\bibliography{2021AAAI}

\begin{thebibliography}{32}
\providecommand{\natexlab}[1]{#1}
\providecommand{\url}[1]{\texttt{#1}}
\providecommand{\urlprefix}{URL }
\expandafter\ifx\csname urlstyle\endcsname\relax
  \providecommand{\doi}[1]{doi:\discretionary{}{}{}#1}\else
  \providecommand{\doi}{doi:\discretionary{}{}{}\begingroup
  \urlstyle{rm}\Url}\fi

\bibitem[{Abadi et~al.(2016)Abadi, Chu, Goodfellow, McMahan, Mironov, Talwar,
  and Zhang}]{abadi2016deep}
Abadi, M.; Chu, A.; Goodfellow, I.; McMahan, H.~B.; Mironov, I.; Talwar, K.;
  and Zhang, L. 2016.
\newblock Deep learning with differential privacy.
\newblock In \emph{Proceedings of the 2016 ACM SIGSAC Conference on Computer
  and Communications Security}, 308--318. ACM.

\bibitem[{Agarwal et~al.(2018)Agarwal, Suresh, Yu, Kumar, and
  McMahan}]{agarwal2018cpsgd}
Agarwal, N.; Suresh, A.~T.; Yu, F. X.~X.; Kumar, S.; and McMahan, B. 2018.
\newblock cpSGD: Communication-efficient and differentially-private distributed
  SGD.
\newblock In \emph{NeurIPS}, 7564--7575.

\bibitem[{Aji and Heafield(2017)}]{aji2017sparse}
Aji, A.~F.; and Heafield, K. 2017.
\newblock Sparse Communication for Distributed Gradient Descent.
\newblock In \emph{Proceedings of the 2017 Conference on Empirical Methods in
  Natural Language Processing}, 440--445.

\bibitem[{Balcer and Cheu(2019)}]{balcer2019separating}
Balcer, V.; and Cheu, A. 2019.
\newblock Separating local \& shuffled differential privacy via histograms.
\newblock \emph{arXiv preprint arXiv:1911.06879} .

\bibitem[{Balle, Barthe, and Gaboardi(2018)}]{balle_privacy_2018}
Balle, B.; Barthe, G.; and Gaboardi, M. 2018.
\newblock Privacy Amplification by Subsampling: Tight Analyses via Couplings
  and Divergences.
\newblock In \emph{Advances in Neural Information Processing Systems},
  volume~31. Curran Associates, Inc.

\bibitem[{Balle et~al.(2019{\natexlab{a}})Balle, Bell, Gascon, and
  Nissim}]{balle_differentially_2019}
Balle, B.; Bell, J.; Gascon, A.; and Nissim, K. 2019{\natexlab{a}}.
\newblock Differentially Private Summation with Multi-Message Shuffling
  \urlprefix\url{http://arxiv.org/abs/1906.09116}.

\bibitem[{Balle et~al.(2019{\natexlab{b}})Balle, Bell, Gasc{\'o}n, and
  Nissim}]{balle_privacy_2019}
Balle, B.; Bell, J.; Gasc{\'o}n, A.; and Nissim, K. 2019{\natexlab{b}}.
\newblock The privacy blanket of the shuffle model.
\newblock In \emph{Annual International Cryptology Conference}, 638--667.
  Springer.

\bibitem[{Bhowmick et~al.(2018)Bhowmick, Duchi, Freudiger, Kapoor, and
  Rogers}]{bhowmick2018protection}
Bhowmick, A.; Duchi, J.; Freudiger, J.; Kapoor, G.; and Rogers, R. 2018.
\newblock Protection against reconstruction and its applications in private
  federated learning.
\newblock \emph{arXiv preprint arXiv:1812.00984} .

\bibitem[{Bittau et~al.(2017)Bittau, Seefeld, Erlingsson, Maniatis, Mironov,
  Raghunathan, Lie, Rudominer, Kode, and Tinnes}]{bittau_prochlo_2017}
Bittau, A.; Seefeld, B.; Erlingsson, Ã.; Maniatis, P.; Mironov, I.;
  Raghunathan, A.; Lie, D.; Rudominer, M.; Kode, U.; and Tinnes, J. 2017.
\newblock Prochlo: Strong Privacy for Analytics in the Crowd.
\newblock In \emph{Proceedings of the 26th Symposium on Operating Systems
  Principles - {SOSP} '17}, 441--459. {ACM} Press.
\newblock ISBN 978-1-4503-5085-3.
\newblock \doi{10.1145/3132747.3132769}.
\newblock \urlprefix\url{http://dl.acm.org/citation.cfm?doid=3132747.3132769}.

\bibitem[{Chan, Shi, and Song(2012)}]{chan2012optimal}
Chan, T.~H.; Shi, E.; and Song, D. 2012.
\newblock Optimal lower bound for differentially private multi-party
  aggregation.
\newblock In \emph{European Symposium on Algorithms}, 277--288. Springer.

\bibitem[{Cheu et~al.(2019)Cheu, Smith, Ullman, Zeber, and
  Zhilyaev}]{cheu_distributed_2019}
Cheu, A.; Smith, A.; Ullman, J.; Zeber, D.; and Zhilyaev, M. 2019.
\newblock Distributed differential privacy via shuffling.
\newblock In \emph{Annual International Conference on the Theory and
  Applications of Cryptographic Techniques}, 375--403. Springer.

\bibitem[{Duchi, Jordan, and Wainwright(2013)}]{duchi2013local}
Duchi, J.~C.; Jordan, M.~I.; and Wainwright, M.~J. 2013.
\newblock Local privacy and statistical minimax rates.
\newblock In \emph{2013 IEEE 54th Annual Symposium on Foundations of Computer
  Science}, 429--438. IEEE.

\bibitem[{Dwork et~al.(2006)Dwork, Kenthapadi, Mcsherry, Mironov, and
  Naor}]{Dwork2006Our}
Dwork, C.; Kenthapadi, K.; Mcsherry, F.; Mironov, I.; and Naor, M. 2006.
\newblock Our Data, Ourselves: Privacy via Distributed Noise Generation.
\newblock In \emph{International Conference on Advances in
  Cryptology-eurocrypt}.

\bibitem[{Dwork, Roth et~al.(2014)}]{dwork2014the(book)}
Dwork, C.; Roth, A.; et~al. 2014.
\newblock The algorithmic foundations of differential privacy.
\newblock \emph{Foundations and Trends in Theoretical Computer Science} 9(3-4):
  211--407.

\bibitem[{Dwork, Rothblum, and Vadhan(2010)}]{dwork2010boosting}
Dwork, C.; Rothblum, G.~N.; and Vadhan, S. 2010.
\newblock Boosting and differential privacy.
\newblock In \emph{2010 IEEE 51st Annual Symposium on Foundations of Computer
  Science}, 51--60. IEEE.

\bibitem[{Erlingsson et~al.(2020)Erlingsson, Feldman, Mironov, Raghunathan,
  Song, Talwar, and Thakurta}]{erlingssonEncodeShuffleAnalyze2020}
Erlingsson, {\'U}.; Feldman, V.; Mironov, I.; Raghunathan, A.; Song, S.;
  Talwar, K.; and Thakurta, A. 2020.
\newblock Encode, {{Shuffle}}, {{Analyze Privacy Revisited}}:
  {{Formalizations}} and {{Empirical Evaluation}}.
\newblock \emph{arXiv:2001.03618 [cs]} 2020arxiv.

\bibitem[{Erlingsson et~al.(2019)Erlingsson, Feldman, Mironov, Raghunathan,
  Talwar, and Thakurta}]{erlingsson_amplification_2018}
Erlingsson, {\'U}.; Feldman, V.; Mironov, I.; Raghunathan, A.; Talwar, K.; and
  Thakurta, A. 2019.
\newblock Amplification by shuffling: From local to central differential
  privacy via anonymity.
\newblock In \emph{Proceedings of the Thirtieth Annual ACM-SIAM Symposium on
  Discrete Algorithms}, 2468--2479. SIAM.

\bibitem[{Geyer, Klein, and Nabi(2017)}]{geyer2017differentially}
Geyer, R.~C.; Klein, T.; and Nabi, M. 2017.
\newblock Differentially private federated learning: A client level
  perspective.
\newblock \emph{arXiv preprint arXiv:1712.07557} .

\bibitem[{Ghazi et~al.(2020)Ghazi, Golowich, Kumar, Manurangsi, Pagh, and
  Velingker}]{ghazi2020pure}
Ghazi, B.; Golowich, N.; Kumar, R.; Manurangsi, P.; Pagh, R.; and Velingker, A.
  2020.
\newblock {Pure Differentially Private Summation from Anonymous Messages}.
\newblock In \emph{1st Conference on Information-Theoretic Cryptography (ITC
  2020)}, volume 163 of \emph{Leibniz International Proceedings in Informatics
  (LIPIcs)}, 15:1--15:23. Dagstuhl, Germany: Schloss Dagstuhl--Leibniz-Zentrum
  f{\"u}r Informatik.

\bibitem[{Ghazi et~al.(2019)Ghazi, Golowich, Kumar, Pagh, and
  Velingker}]{ghazi_power_2020}
Ghazi, B.; Golowich, N.; Kumar, R.; Pagh, R.; and Velingker, A. 2019.
\newblock On the power of multiple anonymous messages.
\newblock \emph{arXiv preprint arXiv:1908.11358} .

\bibitem[{Ghazi, Pagh, and Velingker(2019)}]{ghazi_scalable_2019}
Ghazi, B.; Pagh, R.; and Velingker, A. 2019.
\newblock Scalable and Differentially Private Distributed Aggregation in the
  Shuffled Model \urlprefix\url{http://arxiv.org/abs/1906.08320}.

\bibitem[{Girgis et~al.(2021)Girgis, Data, Diggavi, Kairouz, and
  Suresh}]{girgisShuffledModelFederated2021}
Girgis, A.~M.; Data, D.; Diggavi, S.; Kairouz, P.; and Suresh, A.~T. 2021.
\newblock Shuffled {{Model}} of {{Federated Learning}}: {{Privacy}},
  {{Accuracy}} and {{Communication Trade}}-Offs.
\newblock \emph{IEEE Journal on Selected Areas in Information Theory} 1--1.
\newblock ISSN 2641-8770.
\newblock \doi{10.1109/JSAIT.2021.3056102}.

\bibitem[{Hitaj, Ateniese, and Perez-Cruz(2017)}]{hitaj2017deep}
Hitaj, B.; Ateniese, G.; and Perez-Cruz, F. 2017.
\newblock Deep models under the GAN: information leakage from collaborative
  deep learning.
\newblock In \emph{Proceedings of the 2017 ACM SIGSAC Conference on Computer
  and Communications Security}, 603--618. ACM.

\bibitem[{Liu et~al.(2020)Liu, Cao, Yoshikawa, and Chen}]{liu2020FedSel}
Liu, R.; Cao, Y.; Yoshikawa, M.; and Chen, H. 2020.
\newblock FedSel: Federated SGD under Local Differential Privacy with Top-k
  Dimension Selection.
\newblock In \emph{International Conference on Database Systems for Advanced
  Applications}, 485--501. Springer.

\bibitem[{McMahan et~al.(2016)McMahan, Moore, Ramage, and
  y~Arcas}]{mcmahan2016Federated}
McMahan, H.~B.; Moore, E.; Ramage, D.; and y~Arcas, B.~A. 2016.
\newblock Federated learning of deep networks using model averaging.
\newblock \emph{arXiv preprint arXiv:1602.05629} .

\bibitem[{McMahan et~al.(2018)McMahan, Ramage, Talwar, and
  Zhang}]{mcmahan_learning_2018}
McMahan, H.~B.; Ramage, D.; Talwar, K.; and Zhang, L. 2018.
\newblock Learning Differentially Private Recurrent Language Models.
\newblock In \emph{International Conference on Learning Representations}.

\bibitem[{Nasr, Shokri, and Houmansadr(2019)}]{nasr2019comprehensive}
Nasr, M.; Shokri, R.; and Houmansadr, A. 2019.
\newblock Comprehensive Privacy Analysis of Deep Learning: Passive and Active
  White-box Inference Attacks against Centralized and Federated Learning.
\newblock In \emph{2019 {IEEE} Symposium on Security and Privacy, {SP} 2019,
  San Francisco, CA, USA, May 19-23, 2019}, 739--753. {IEEE}.

\bibitem[{Shokri et~al.(2017)Shokri, Stronati, Song, and
  Shmatikov}]{shokri2017membership}
Shokri, R.; Stronati, M.; Song, C.; and Shmatikov, V. 2017.
\newblock Membership inference attacks against machine learning models.
\newblock In \emph{2017 IEEE Symposium on Security and Privacy (SP)}, 3--18.
  IEEE.

\bibitem[{Wang et~al.(2019{\natexlab{a}})Wang, Xiao, Yang, Zhao, Hui, Shin,
  Shin, and Yu}]{wang2019collecting}
Wang, N.; Xiao, X.; Yang, Y.; Zhao, J.; Hui, S.~C.; Shin, H.; Shin, J.; and Yu,
  G. 2019{\natexlab{a}}.
\newblock Collecting and Analyzing Multidimensional Data with Local
  Differential Privacy.
\newblock In \emph{2019 IEEE 35th ICDE}, 638--649.

\bibitem[{Wang et~al.(2019{\natexlab{b}})Wang, Xu, Ding, Zhou, Hong, Huang, Li,
  and Jha}]{wang2019murs}
Wang, T.; Xu, M.; Ding, B.; Zhou, J.; Hong, C.; Huang, Z.; Li, N.; and Jha, S.
  2019{\natexlab{b}}.
\newblock MURS: Practical and Robust Privacy Amplification with Multi-Party
  Differential Privacy.
\newblock \emph{arXiv} arXiv--1908.

\bibitem[{Yang et~al.(2019)Yang, Liu, Chen, and Tong}]{yang2019Federated}
Yang, Q.; Liu, Y.; Chen, T.; and Tong, Y. 2019.
\newblock Federated Machine Learning: Concept and Applications.
\newblock \emph{ACM Transactions on Intelligent Systems and Technology} 10(2):
  1--19.

\bibitem[{Zhu, Liu, and Han(2019)}]{zhu2019deep}
Zhu, L.; Liu, Z.; and Han, S. 2019.
\newblock Deep Leakage from Gradients.
\newblock In \emph{Advances in Neural Information Processing Systems},
  volume~32. Curran Associates, Inc.

\end{thebibliography}

\clearpage
	
\end{document}